\documentclass[journal,onecolumn]{IEEEtran}
\usepackage{graphicx}
\usepackage{amsmath}
\usepackage{amsfonts}
\usepackage{subfig}
\usepackage{cite}
\usepackage{amsmath}
\usepackage{algorithm}
\usepackage[noend]{algpseudocode}
\usepackage{subfig}
\usepackage{float}
\usepackage{array}
\usepackage{caption}
\usepackage[table]{xcolor}
\usepackage{booktabs}
\newcolumntype{P}[1]{>{\centering\arraybackslash}p{#1}}
\newcolumntype{M}[1]{>{\centering\arraybackslash}m{#1}}

\algrenewcommand\algorithmicforall{\textbf{foreach}}
\algrenewcommand\algorithmicindent{.8em}

\ifCLASSINFOpdf
\else
\fi

\begin{document}
\title{An Interval Type-2 Fuzzy Approach to Automatic PDF Generation for Histogram Specification}

\author{Vishal Agarwal*, Diwanshu Jain*, A. Vamshi Krishna Reddy*, 
        and Frank Chung-Hoon Rhee,~\IEEEmembership{Member,~IEEE}% <-this % stops a space
\thanks{D. Jain, A. V. K. Reddy and V. Agarwal are with the Department
of Electronics and Electrical Engineering, Indian Institute of Technology Guwahati (IITG), India. Email: \{diwanshu, vamshi, vishal.agarwal\}@iitg.ernet.in }% <-this % stops a space
\thanks{F. C.-H. Rhee is with Hanyang University, Korea. % <-this % stops a space
Email: frhee@fuzzy.hanyang.ac.kr}
\thanks{The project was conceptualized and completed at the Computational Vision
and Fuzzy Systems Lab, Hanyang University.}
\thanks{* represents equal contribution.}
}

\maketitle

\begin{abstract}
Image enhancement plays an important role in several application in the field of computer vision and image processing. Histogram specification (HS) is one of the most widely used techniques for contrast enhancement of an image, which requires an appropriate probability density function for the transformation. In this paper, we propose a fuzzy method to find a suitable PDF automatically for histogram specification using interval type - 2 (IT2) fuzzy approach, based on the fuzzy membership values obtained from the histogram of input image. The proposed algorithm works in 5 stages which includes - symmetric Gaussian fitting on the histogram, extraction of  IT2 fuzzy membership functions (MFs) and therefore, footprint of uncertainty (FOU), obtaining membership value (MV), generating PDF and application of HS. We have proposed 4 different methods to find membership values - point-wise method, center of weight method, area method, and karnik-mendel (KM) method. The framework is sensitive to local variations in the histogram and chooses the best PDF so as to improve contrast enhancement. Experimental validity of the methods used is illustrated by qualitative and quantitative analysis on several images using the image quality index - Average Information Content (AIC) or Entropy, and by comparison with the commonly used algorithms such as Histogram Equalization (HE), Recursive Mean-Separate Histogram Equalization (RMSHE) and Brightness Preserving Fuzzy Histogram Equalization (BPFHE). It has been found out that on an average, our algorithm improves the AIC index by $11.5 \%$ as compared to the index obtained by histogram equalisation.
\end{abstract}

\begin{IEEEkeywords}
Automatic PDF generation, Fuzzy sets, Contrast enhancement, Histogram Specification, Image enhancement, Interval Type-2 fuzzy
\end{IEEEkeywords}

% \fi
\IEEEpeerreviewmaketitle

\section{Introduction}
\IEEEPARstart{I}{mage} enhancement is one of the important domain of image processing which aims to improve the appearance of the image for display or further analysis. It is an important preprocessing step in most computer vision and image processing pipelines. Computer Vision aims to automate the process of visual perception. It enhances a vision system's ability to detect objects, resulting in an overall improvement in accuracy. Image enhancement helps us to transform image on the basis of psychophysical characteristics of human visual system\cite{i1}. The different enhancement algorithms may be classified according to the objective achieved, such as filtering, noise removal, edge and contrast enhancement. In particular, contrast manipulation techniques are applied on the images to extract information which are not easily distinguished on low contrast images. It is widely used to achieve wider dynamic range\cite{i2}.
 
Histogram specification and Histogram equalization are two of the most well-known algorithms, in the class of contrast enhancement, due to its simplicity and effectiveness\cite{i3}. Histogram specification (HS) redistribute the gray levels of an input image such that the distribution of the transformed gray levels closely matches a given desirable probability density function (PDF). Histogram equalization (HE) is a special case of HS in which the desired PDF is an uniform distribution. HE results in improved dynamic range of the image but conventional HE is rarely used in consumer electronic products such as TV, mobile phones, due to its inability to preserve brightness\cite{i4}.
 
Several algorithms have been proposed to overcome HE’s disadvantages\cite{i5,i6,i7,i8,i9,i10,i11,i12,i13}. The bi-histogram equalization (BBHE) is a part of class of algorithms which aims at preserving the mean brightness of a histogram-equalized image\cite{i4}. BBHE decomposes an image into two sub-images based on the mean brightness of the image and performs HE on each sub-image\cite{i9}. Many algorithms have been proposed to improve upon BBHE such as dualistic sub-image histogram equalization (DSIHE), minimum mean brightness error bi-histogram equalization (MMBEBHE) and recursive mean-separate histogram equalization (RMSHE)\cite{i10,i11,i12,i13}. In general, algorithms in this class do not modify the gray level transformation function but instead modify the region(s) of application of HE.
 
Fuzzy logic has found many applications in image processing, pattern recognition, etc. Image associated with imprecision can  be handled efficiently by application of Fuzzy set theory. Some fuzzy techniques have also been proposed for contrast enhancement\cite{i14,i15,i16,i17,i18}. Fuzzy histogram equalization (FHE) extends traditional HE by using a transformation based on fuzzy weighted values assigned to each gray level\cite{i15}. In\cite{i16}, the gray levels are transformed according to a parametrised fuzzy membership function (MF). However, the MFs take on a fixed form and hence may not be appropriate for all images. In\cite{i17}, the desired histogram is modelled as one parametrised fuzzy number and fuzzy inference rules and operations are used to tune this number and to decide the slope of mapping function. Although the algorithm is automatic, tuning of the fuzzy number should be performed carefully for good generalization. In\cite{i18}, a fuzzy algorithm was proposed which considered both global and local information of the image using the fuzzy entropy principle.
 
In conventional histogram specification, the primary difficulty is to get a meaningful and appropriate transformation analytically. This ambiguity may be resolved by choosing a set of functions which are considered suitable, and providing hyper parameters which improve the generality of the algorithm. In\cite{i19,i20}, modified cosine functions and genetic algorithms are used to find suitable transform functions. In\cite{i21}, the cumulative density function (CDF) of a given image is approximated as a piecewise-linear function and mapped to a PDF for usage in HS. Dynamic histogram specification (DHS) proposed in\cite{i22} generates an appropriate PDF using gray levels (critical points) in the histogram whose first derivative is greater than a specific gain value and whose second derivative is equal to zero. DHS may enhance image contrast without losing the shape features of original histogram. \par

Type-2 fuzzy sets are finding very wide applicability in rule-based fuzzy logic systems (FLSs) because they let uncertainties be modelled by them whereas such uncertainties cannot be modelled by type-1 fuzzy sets. Type-1 fuzzy, despite having a name which carries the connotation of uncertainty, research has shown that there are limitations in the ability of T1 FSs to model and minimize the effect of uncertainties \cite{i23,i24,i25}. This is because a T1 FS is certain in the sense that its membership grades are crisp values.  Recently, type-2 FSs\cite{i26}, characterized by MFs that are themselves fuzzy, have been attracting interests. Interval type-2 (IT2) FSs1, a special case of type-2 FSs, are currently the most widely used for their reduced computational cost. We use the concept of IT2 fuzzy to estimate the appropriate PDF for a given image. The selection of PDF occurs automatic in contrast to the conventional histogram specification where we need to provide the PDF.
 
In this paper, we propose a simple fuzzy approach for automatic generation of a suitable PDF from the input image for contrast enhancement using HS. The proposed algorithm works in 5 stages - symmetric Gaussian fitting on the input histogram, generation of IT2 fuzzy MF and FOU, obtaining membership value (MV), PDF generation, and histogram specification. The authentication of result is done by using average information content(AIC) index. 

The rest of the paper follows the following organisation- Section II provides the pre-requisites for the research which includes a brief introduction to fuzzy sets, histogram equalisation and histogram specification. Section III provides the detailed description of the proposed algorithm, and a brief discussion of its salient features in 5 sub-sections. Experimental results and discussion can be found in Section IV. Lastly, Section V presents the conclusion of our work and further research possibilities.

\section{Background}

The concepts which are extensively used in the paper are: Fuzzy Sets, Histogram Equalization and Histogram Specifications. These fundamental building blocks are briefly described as follows:
\subsection{Type-1 and Type-2 Fuzzy sets}
In classical set theory, membership of an element $x$ belonging to a domain of discourse (universe) $U$ to a set $A$ may be represented as a binary function $\mathrm{\mu}_A(x)$ which is defined as in \eqref{first}

\begin{eqnarray} \label{first}
\mu_A(x) = \begin{cases}
           1  & \text{if } x \in A \\ \\
           0  & \text{if } x \not\in A  
       \end{cases} 
\end{eqnarray}

where $\mu_A(x)$ is called the \textit{membership function}. Set $A$ (which can also be treated as a subset of $U$) is mathematically equivalent to its membership function $\mu_A(x)$ in the sense that knowing $\mu_A(x)$ is the same as knowing $A$ itself.\cite{b1,b2}\par
\begin{figure}[H]
\centering
\subfloat[]{\includegraphics[scale = .35]{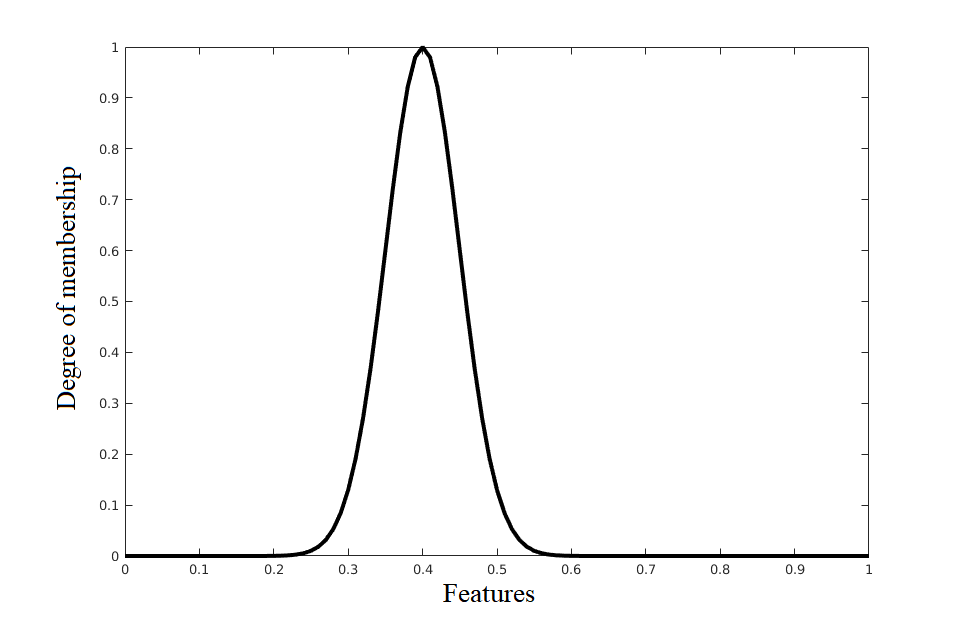}
\label{A T1 MF}}
\hfil
\subfloat[]{\includegraphics[scale = .35]{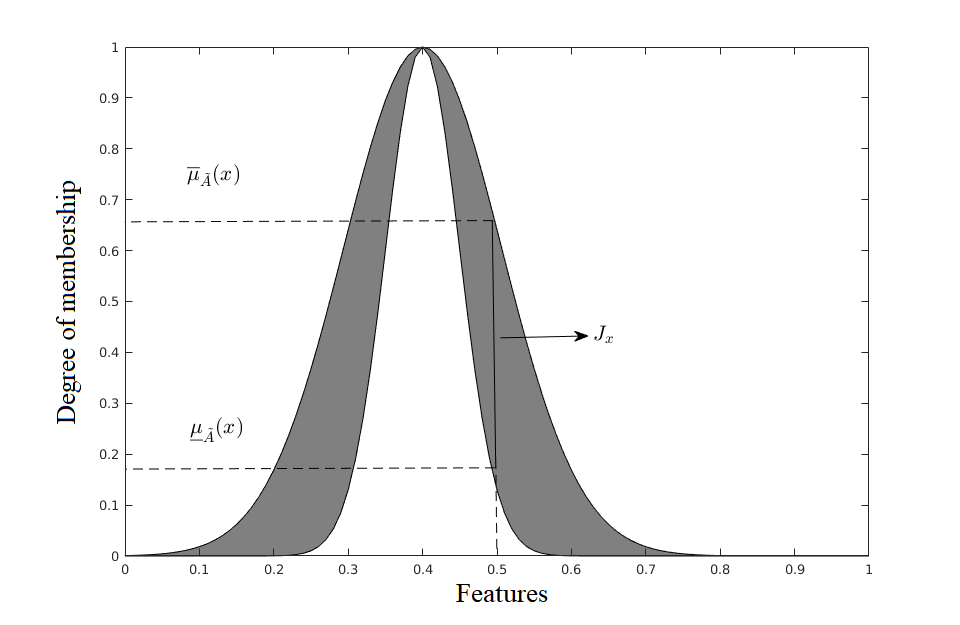}
\label{A T2 MF}}
\caption{Membership function (a) Type-1, (b) Type-2}
\label{fig_sim1}
\end{figure}

Zadeh\cite{b3} proposed a non-binary, function-theoretic representation of set membership as a mechanism for handling uncertainty. The membership function for such fuzzy sets represents the degree of membership of an element to the set.  A Type-1 Fuzzy Set (T1 FS) $A$ is comprised of domain $U$ of the real numbers (called the \textit{universe of discourse} of A) together with the \textit{membership function} (MF) $\mu_A):U \to[0,1]$. For each value of $x$, $\mu_A(x)$ is the \textit{degree of membership} or \textit{membership grade}, of $x$ in $A$. When $U$ is continuous, $A$ is written as \begin{eqnarray}
A = \int_U \mu_A(x)/x
\end{eqnarray}
fig. \ref{fig_sim1}\subref{A T1 MF} shows an example of type-1 fuzzy membership function. Detailed discussion of fuzzy sets and operations can be found in\cite{b2}.\par
A type-2 fuzzy set (T2 FS) $\widetilde{A}$ is the extension of the concept of T1 FS. It introduce the concept of uncertainty in membership grades of T1 FSs. Mathematically, A \textit{T2 FS} (also called a \textit{GT2 FS}) is denoted by a bivariate function on the Cartesian product $\mu:X\times[0,1] $ into $[0,1]$, where $X$ is the universe of discourse for the \textit{primary variable} of  $\widetilde{A},\,x$. The 3D MF of  $\widetilde{A}$ is usually denoted as $\mu_{\widetilde{A}}(x,u)$, where $x\in U$ and $u\in U =[0,1]$, that is \begin{eqnarray}
\widetilde{A} = {((x,u),\mu_{\widetilde{A}}) | \forall x\in X, {\forall u\in J_x} \subseteq[0,1]}
\end{eqnarray} 
in which $0\leq\mu_{\widetilde{A}}\leq 1$. $\widetilde{A}$ can also be expressed as \begin{eqnarray}
\widetilde{A} = \int_{x\in X}\int_{u \in J_x}\mu_{\widetilde{A}}(x,u)/(x,u)\,\,where \,\,J_x \subseteq[0,1]\end{eqnarray} where
$\int\int$ denotes union over all admissible $x$ and $u$ and $J_x$ is the range of primary membership. For discrete universes of discourses $\int$, $X$ and $U$ is replaced by $\Sigma$, $X_d$ and $U_d$ respectively. Fig.\ref{fig_sim1}\subref{A T2 MF} represents a graphical representation of a T2 MF. A vertical slice of the T2 MF at $x = x'$ defines the secondary membership function at that point, given by\cite{b2}
\begin{eqnarray}
\mu_{\widetilde{A}} = f_{x'}(u) = {\int_{u \in J_x'}\frac{\mu_{\widetilde{A}}(x',u)}{u}\, ,where \,\,  J_x \subseteq[0,1]}
\end{eqnarray}

Uncertainty in the primary membership of a T2 FS is termed as \textit{footprint of uncertainty} (FOU) and the bounds of the FOU are accordingly called the upper and lower MFs. A FS which does not impose any additional constraints on the secondary membership function is known as a \textit{general T2 FS}. One class of MFs that have gained immense popularity due to its mathematical traceability and ease of computability are \textit{interval Type-2 FS} (IT2). The secondary MFs of IT2 Fss are interval sets represented as
\begin{eqnarray}
\widetilde{A} = \int_{x\in X} \frac{\int_{u\in J_x} \frac{1}{u}}{x} \,\,\, where J_x \subseteq[0,1]\end{eqnarray}

\begin{figure}
\centering
\includegraphics[scale=.65	]{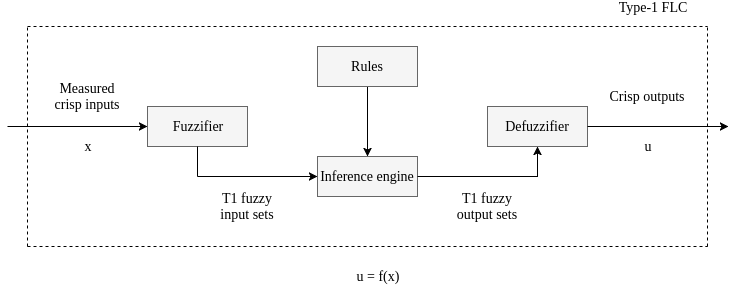}
\caption{{General Structure of a type-1 FLC}}\label{General Structure of a type-1 FLC}
\end{figure}

A fuzzy logic system (FLS) or a fuzzy inference system defines a mapping of input data to a scalar valued output using fuzzy rules\cite{b1}. Typically, a FLS consists of four major components - a fuzzifier, a fuzzy rule base, an inference engine, and a defuzzifier\cite{b2}. Fig. \ref{General Structure of a type-1 FLC} shows the block diagram for T1 FLS. The non-fuzzy input values are converted to fuzzy variables by the fuzzifier and processed by an inference engine using the fuzzy rule base and then converted back to crisp output value(s) using the defuzzifier. The defuzzifier may contain a \textit{type-reducer} for the conversion of high dimensional fuzzy sets (T2 or above) to T1 sets. Computing centroid of the set is generally performed for type reduction of T2 fuzzy set. Karnik and Mendel\cite{b2} have developed iterative algorithm (known as \textit{KM algorithm}) for computing centroid. Easier methods like \textit{mean value computation of the FOU of primary MF} are available for IT2. \par

Type-1 FLSs have been developed and successfully implemented in different areas of engineering. For example, in linguistic evaluation of machine tools, modelling different operational research problems and fault detection in gearboxes\cite{b4}. \par
T1 FSs are crisp and precise in a sense that there MFs are known perfectly, which is not usually the case. T2 FSs adds a layer of uncertainty by introducing uncertainty in MFs i.e. \textit{fuzzy MFs}. MFs are 3-Dimension in this case. Shaded region in Fig. \ref{fig_sim1}\subref{A T2 MF} shows the cross section of FOU of a T2 MF. T2 FSs have been shown to be more effective than T1 FSs in the connection admission control (CAC) method in ATM networks\cite{b5}, prediction of Mackey-Glass chaotic time-series\cite{b6}, adaptive noise cancellation of signals\cite{b7}, and in modelling of radiographic tibia image data using neuro-fuzzy clustering\cite{b8}, among several other applications.

\subsection{Histogram Equalization}
Histogram Equalization (HE) is a technique for adjusting image intensities to enhance contrast.
Let $I(x,y)$ be a given image as a matrix of integer pixel intensities ranging from $0$ to $L-1$. $L$ is the number of possible intensity values (gray levels), $256$ in our case. Let $p_n$ denote the normalized histogram of $I(x,y)$ with a bin for each possible intensity. So, \begin{eqnarray}
p_n = \frac{no.\,of\,pixels\,with\,intensity\,n}{total\,no.\,of\,pixels}\,\,\,\,\,\,\,n = 0,1,\cdots,L-1 \end{eqnarray}
The probability of occurrence of a gray level in the input image may be approximated by the PDF  $P_I(g)$ as \begin{eqnarray} P_I(g) = \frac{n(g)}{n},\,\,\,\, g\in {g_0,\cdots,g_{(L-1)}}
\end{eqnarray}
where $g$ is a gray level, $n(g)$ is the number of pixels with gray level equal to $g$ and $n$ is total number of pixels in the image[3]. In HE, the transformation function $T(g_i)$ mapping an input gray level $g_i$ to an output level $g_f$ is given by \begin{eqnarray}\label{dynamicrange}
g_f = T(g_i)= \displaystyle{\sum\limits_{g \leq g_i} \frac{n(g)}{n}}
\end{eqnarray}
This transformation is also known as \textit{histogram linearisation}\cite{i3}, and is mathematically equivalent to \textit{cumulative frequency histogram}. It has been shown that in the case of continuous gray level values, the continuous linearisation of input gray levels produces a uniform probability distribution of output gray levels, irrespective of the form of $P_I(g)$\cite{i3}. It has been observed the application of \eqref{dynamicrange} increases the dynamic range of output image histogram by \textit{flattening} the histogram\cite{i4}. \par
There are following disadvantages in using HE-
\begin{itemize}
\item Brightness Alteration
\item Detail Loss
\item Can produce undesirable effects when applied to low colour depth
\item Is indiscriminate. It may increase the contrast of background noise, while decreasing the usable signal.			%REF WIKI
\end{itemize}

% FIGURE FOR SECTION III

\subsection{Histogram Specification}
Histogram Specification (HS) is a technique that transforms gray levels of an input image $I(x,y)$ such that the histogram of transformed image approximates a given PDF. HE is the special case of HS where PDF used is the constant PDF. For the desired $P_D(g)$, we define the following transformations: \begin{eqnarray}\label{trans9}
G(g_f) = \displaystyle{\sum\limits_{g \leq g_f}P_D(g)}
\end{eqnarray}\begin{eqnarray}\label{trans10}
T(g_i) = \displaystyle{\sum\limits_{g \leq g_i}\frac{n(g)}{n}}
\end{eqnarray}
where $g_f$ represents a gray level in the output image and $g_i$ represents a gray level in the input image. All other symbols have the same meaning as before. Mapping between the output levels and the input levels is given by the transformation-\begin{eqnarray}\label{trans11}
g_f = G^{-1}(T(g_i))
\end{eqnarray}\par
The procedure of HS can be summarized as follows-
\begin{enumerate}
\item Using the desired PDF, obtain the transformation function $G(g_f)$ using \eqref{trans9}
\item Equalize the gray levels of the original image using \eqref{trans10}
\item Using \eqref{trans11}, map the input gray levels to a suitable output gray level.
\end{enumerate}
\par
Due to strong dependence of output image on the supplied PDF, it's crucial to select the appropriate PDF so as to improve the contrast enhancement.

\section{Proposed Method: Automatic Fuzzy Histogram Specification using IT2 FS}

The proposed algorithm consists of 5 stages: symmetric Gaussian fitting on the histogram, generation of IT2 fuzzy MF and hence Footprint of Uncertainty (FOU), obtaining Membership Value (MV), generation of desired PDF and transformation of histogram of input image to that of output image, using Histogram Specification. The rest of the section explains the five stages in detail followed by the precise summary of our proposed method.
\subsection{Symmetric Gaussian fitting on the input histogram}
The first stage has following key steps to be done:\\

	\subsubsection{Input Histogram preprocessing}
	The histogram of an image is smoothed using a moving average filter. Another methods available for smoothing are using a symmetric window such as triangular window or a hyper-cube or to fit data with a smoothing spline[32]. The result is then normalized and is used in further processing. Fig. \ref{gfit}\subref{HistogramOrgImage} shows the initial histogram of image given in Fig. \ref{gfit}\subref{OriginalImage} and the smoothed and normalized histogram is shown in Fig. \ref{gfit}\subref{Smooth Normal Histogram}\\

	\subsubsection{Gaussian fitting}
A general Gaussian is given by \eqref{generalgaussian}
\begin{eqnarray} \label{generalgaussian}
G(\textbf{x}) = a\cdot exp\left(-\frac{1}{2}(\textbf{x} - \textbf{$\mu$}^T)\Sigma^{-1}(\textbf{x}-\textbf{$\mu$})\right)
\end{eqnarray}
where $a$ is the height, $\mu$ is the mean vector, and $\Sigma$ is the covariance matrix. We can model a histogram having $n$ peaks as the sum of Gaussian functions as \begin{eqnarray}
\widetilde{G}(\textbf{x}) = \sum\limits_{i = 1}^{n}G_i(\textbf{x})
\end{eqnarray}
To approximate the smoothed histogram as Gaussian functions, the objective function to be minimised is \begin{eqnarray}
J(\textbf{p}_i) = \frac{1}{2}\left(\sum\limits_{i=1}^n G_i(\textbf{x}) - H(\textbf{x})\right)^2
\end{eqnarray}
	where parameter vector $\textbf{p} = (a_i,\mu_i,\Sigma_i)$ is for the $i^{th}$ Gaussian function $G_i(\textbf{x})$ and $H(\textbf{x})$ is the smoothed histogram of the input data. We can apply gradient descent method to estimate the parameter vector $\textbf{p}_i$ which uses the following update rule- \begin{eqnarray}
	{\textbf{p}_i}^{new} = {\textbf{p}_i}^{old} -\rho\frac{\partial J}{\partial\textbf{p}_i}
	\end{eqnarray}
	where $\rho$ is a positive learning constant.\cite{p1}\par
	For Gaussian functions, the partial derivatives of $J$ with respect to each component of $\textbf{p}_i$ are \begin{eqnarray}
	\frac{\partial J}{\partial a_i} = \left(\sum\limits_{j=}^n G_j(\textbf{x}) - H(\textbf{x})\right)\cdot exp\left(-\frac{1}{2}(\textbf{x}-\textbf{$\mu$}_i)^T{\Sigma^{-1}}_i (\textbf{x} - \textbf{$\mu$}_i)\right)
	\end{eqnarray}
	\begin{eqnarray}
	\frac{\partial J}{\partial \textbf{$\mu$}_i} = \frac{1}{2}\left(\sum\limits_{j=1}^n G_j(\textbf{x})-H(\textbf{x})\right)\cdot G_i(\textbf{x})\cdot(\textbf{x}-{\textbf{$\mu$}_i})^T\cdot({\Sigma_i}^{-T} + {\Sigma_i}^{-1})
	\end{eqnarray}
	\begin{eqnarray}
	\frac{\partial J}{\partial \textbf{$\Sigma$}_i} = \frac{1}{2}\left(\sum\limits_{j=1}^n G_j(\textbf{x})-H(\textbf{x})\right)\cdot G_i(\textbf{x})\cdot\left({\textbf{$\Sigma$}_i}^{-T}(\textbf{x}-\textbf{$\mu$}_i)\cdot{(\textbf{x}-\textbf{$\mu$}_i)}^T{\textbf{$\Sigma$}_i}^{-T}\right)
	\end{eqnarray}
	
\begin{figure}[!t]
\centering
\subfloat[]{\includegraphics[scale = .33]{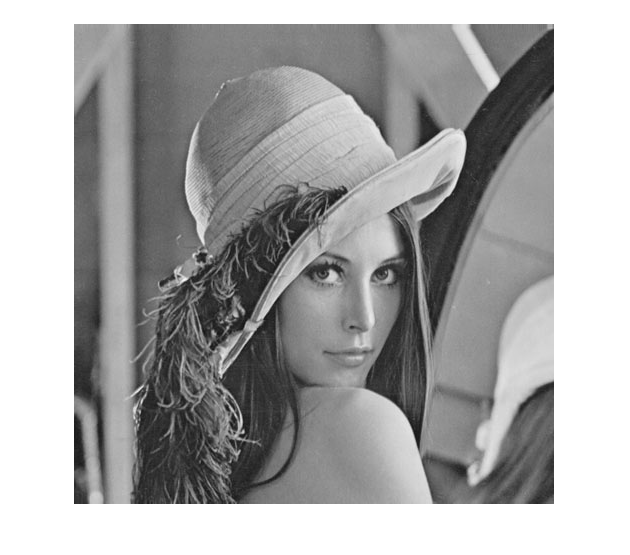}
\label{OriginalImage}}
\subfloat[]{\includegraphics[scale = .45]{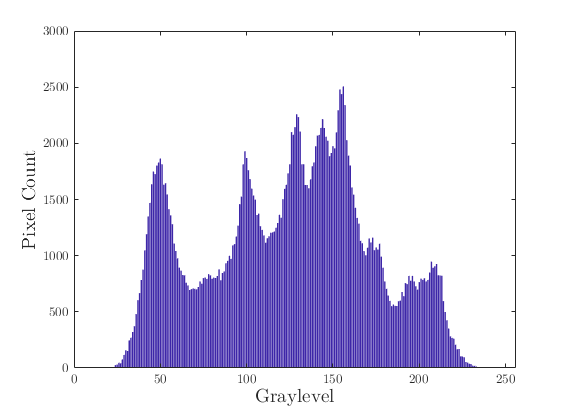}
\label{HistogramOrgImage}}
\hfil
\subfloat[]{\includegraphics[scale = .45]{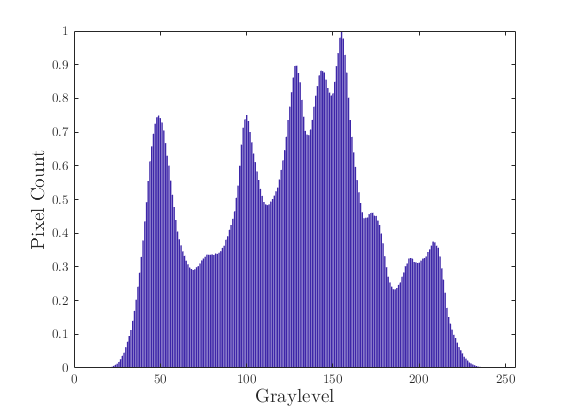}
\label{Smooth Normal Histogram}}
\subfloat[]{\includegraphics[scale = .45]{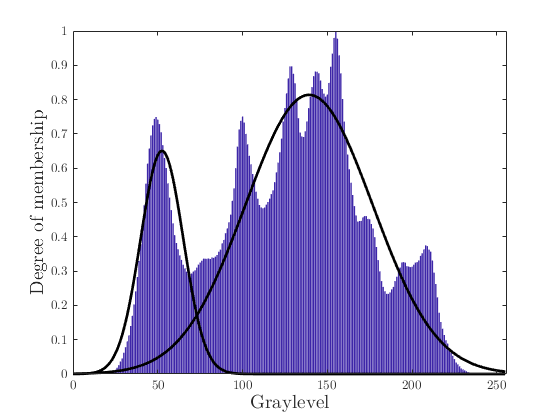}
\label{gaussfit}}
\caption{Illustration of gaussian fitting : (a) input image - \lq{Lena}\rq, (b) histogram of input image, (c) smoothened and normalized histogram, (d) best-fit gaussian}
\label{gfit}
\end{figure}

	For gradient descent methods, choice of the initial values for the parameters is critical, due to the local minima problem. Therefore, we can use the following heuristic approach that consists of following steps to obtain the initial parameters.\cite{p1}
	\begin{enumerate}
	\item Using the least squares approximation, fit a polynomial function (PF) of the lowest possible degree (i.e. to avoid over fitting) such that the fit to each smoothed histogram has a reasonably small error.
	\item Calculate the extrema (maxima and minima) values for the PF in Step 1 and determine the the position of the all Gaussians by the observing the positive values maxima, ignoring the ones that have small peaks.
	\item Initialise the heights of the Gaussians by the maxima values (peak values) and initialise the mean values of the Gaussians as the location of these peaks. Initialise the standard deviation of each Gaussian as the shortest value among the distances between the mean of the Gaussian and the nearest minima or roots of the PF.
	\item The mid-value of the two identified consecutive peak(s) are considered to be the partition point(s) ${pp}_i$ which is used further in determining membership values.
	\end{enumerate}

Fig. \ref{gfit}\subref{gaussfit} shows the Gaussian fitting over the given histogram.

\subsection{Generation of IT2 fuzzy MF and FOU}
The values of the histogram which lie above and below the best-fit curve $(G(g))$ are identified. The values exceeding this curve are used to generate the upper MF (by the process of Gaussian fitting) and analogously, the values below the curve are used to generate the lower MF. For LMF, Gaussian is fitted over the function $L(g)$ which is obtained by \eqref{L(g)} -
\begin{eqnarray} \label{L(g)}
L(g) = min\, \{G(g),H(g)\} =  
\begin{cases}
           H(g)  & \text{if } H(g) < G(g) \\ \\
           G(g)  & \text{if } H(g) \geq G(g)  
       \end{cases} 
\end{eqnarray}
Similarly, For UMF, a function $U(g)$ can be defined over which Gaussian is fitted using \eqref{U(g)}\begin{eqnarray} \label{U(g)}
U(g) = max\, \{G(g),H(g)\} = \begin{cases}
           G(g)  & \text{if } H(g) \leq G(g)  \\ \\
           H(g)  & \text{if } H(g) > G(g)
       \end{cases} 
\end{eqnarray}

\begin{figure}[H]
\centering
\subfloat[]{\includegraphics[scale = .45]{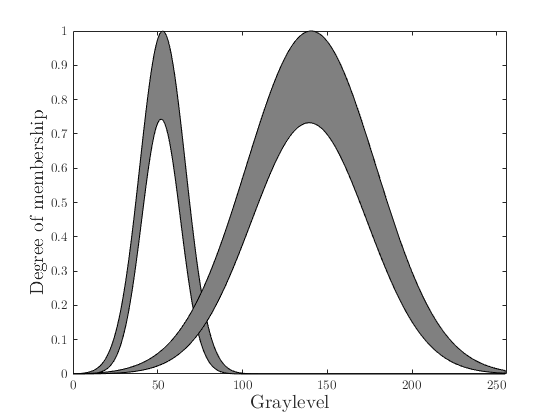}
\label{fou}}
\hfil
\subfloat[]{\includegraphics[scale = .35]{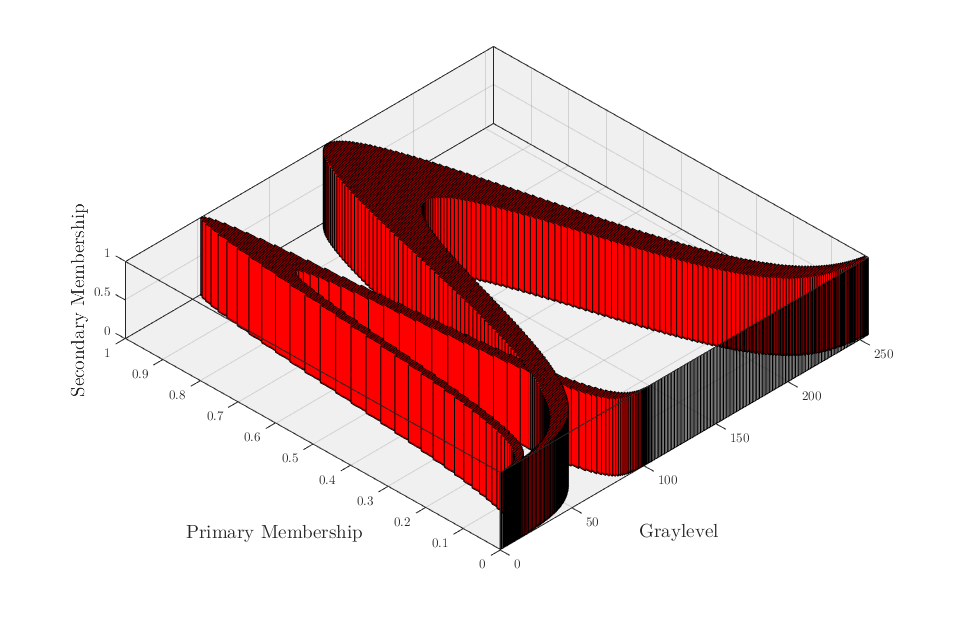}
\label{FOU3d}}
\caption{Illustration of IT2 membership function : (a) footprint of uncertainty (FOU) and (b) the fuzzy MF}
\label{fig_simfou}
\end{figure}

The upper and lower MF are modelled as sum of Gaussian Functions. This sum may be expressed as \begin{eqnarray} \label{F_tilda}
F\widetilde(x) = \sum\limits_{i=1}^N F_i(x),
\end{eqnarray}
where $F_i(x)$ is defined as in \eqref{eq of gaussian}, $N$ is the number of functions. \par
\begin{eqnarray}\label{eq of gaussian}
F(x) = a\cdot exp\left(-\frac{1}{2}\frac{(x-\mu)^2}{{\sigma}^2}\right)
\end{eqnarray}

where $a$ is the height, $\mu$ is the center, and $\sigma$ is the standard deviation of the Gaussian. Thus, the upper MF (UMF) represents an upper bound functional approximation of the image histogram while the lower MF (LMF) represents a lower bound approximation. Fig.    \ref{fig_simfou}\subref{fou} shows the obtained footprint of uncertainty.\par

\begin{figure}[H]
\centering
\subfloat[]{\includegraphics[scale = .45]{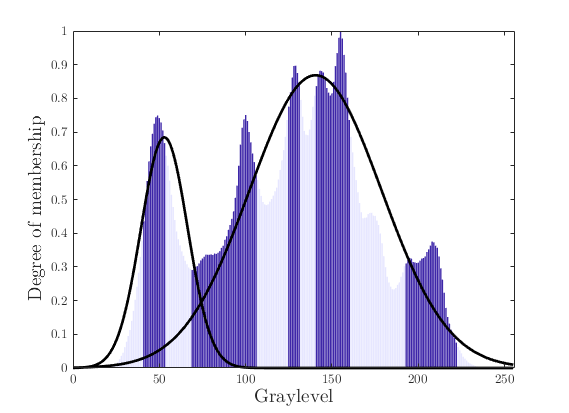}
\label{fitumf}}
\hfil
\subfloat[]{\includegraphics[scale = .45]{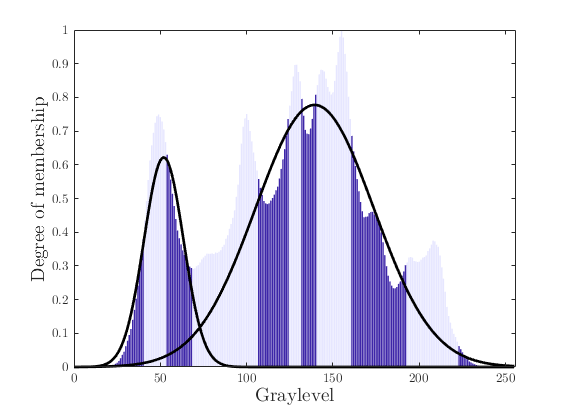}
\label{fitlmf}}
\caption{Illustration of gaussian fitting for (a) UMF and (b) LMF}
\label{fig_sim12}
\end{figure}

We define the \textit{reach} of a function $F_i(x)$ as the interval in which the value attained by $F_i(x)$ is greater than the value of all the other symmetric Gaussian functions. Mathematically, the \textit{reach} of the function $F_i(x)$ is the closed interval $[c_{i_1},c_{i_2}]$ such that \begin{eqnarray}\label{reach}
F_i(x) \geq F_j(x),  \quad \forall x\in [c_{i_1},c_{i_2}],\quad j\in \{1,\cdots,N\}.
\end{eqnarray}
where the start and end points of each interval $\left(c_{i_1}\right.$ and $\left.c_{i_2}\right)$ are called the \textit{crossover} values of the function. From \eqref{reach}, it can be deduced that two symmetric Gaussian functions defining the histogram can have overlapping reach only at the crossover points.\par

We define the \textit{domain} of a gray level $d(g)$ as 
\begin{eqnarray}\label{domain}
d(g) = min\,\{i|g \in [c_{i_1},c_{i_2}]\}
\end{eqnarray} 
Thus, \textit{domain} refers to a function in whose reach a particular gray level lies. For levels which may lie in multiple reaches, we arbitrarily choose the smallest function whose reach contains that point. The domain at crossover point is taken as the minimum of the possible values. The notions of reach, crossover points and domain is used in PDF generation in the upcoming section. \par
The sections obtained after Gaussian fitting can be assigned a fuzzy label. For example, for 2 Gaussian fitting, it can be classified as dark and bright, and hence clustering of gray levels can be done efficiently.  
 
\subsection{Obtaining Membership Value (MV)}
A fuzzy \textit{Membership Value} (MV) is defined as a transformation from a fuzzy MF to a real number. Since the proposed algorithm utilizes an IT2 MF, MV computations is performed on the upper and the lower MFs using piecewise linear or affine functions, thereby saving computational expense which is incurred in computation for a general T2 MF. That is, IT2 MV of a gray level is a pair of real numbers obtained by sequentially applying a linear or affine MV computation technique to the upper and lower MFs extracted in the previous stage. The manner of formulation of the computation technique affects the manner in which the PDF is generated, which can be seen from \eqref{upperPDF}, \eqref{lowerPDF} and \eqref{PDFfinal}. Four possible methods for membership value computation are as follows:\\
\subsubsection{Fuzzy Membership Value- Point-wise Method}

\begin{figure}[H]
\centering
\subfloat[]{\includegraphics[scale = .45]{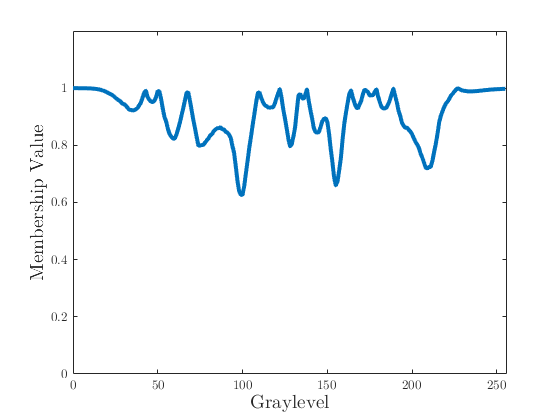}
\label{PW Upper MV}}
\hfil
\subfloat[]{\includegraphics[scale = .45]{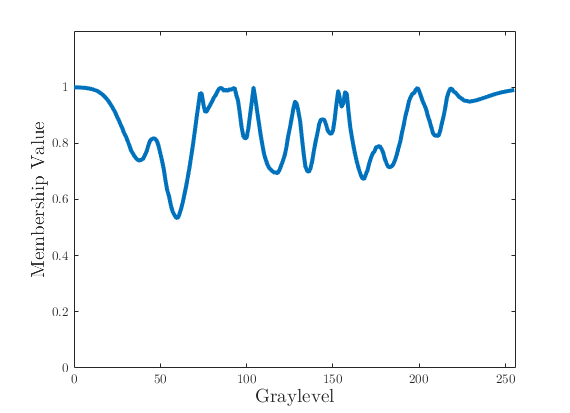}
\label{Lower MV}}
\caption{Memebership Value using point-wise method : (a) upper MV and (b) lower MV}
\label{fig_simPW}
\end{figure}

It is based on the difference in the values of the fuzzy MF (lower/upper) and the histogram for each gray level. The membership value of a gray level $g$ with respect to the $i^{th}$ membership function $F_i$ is given by 
\begin{eqnarray}\label{pointwise mv1}
M_{PW}(g,i) = 1-|F_i(g)-H(g)|
\end{eqnarray}
where $H(g)$ is the histogram value. Thus, the membership value of a gray level decreases as the difference between the fuzzy MF (histogram approximation) and the histogram increase. The PDF generation technique takes this into consideration by assigning higher probabilities to smaller values, which leads to better distribution of gray levels in the output histogram. The overall membership value $M_v(g)$ for a particular MF(upper/lower) of a gray level $g$ is given by \begin{eqnarray}\label{pointwise mv2}
M_v(g) = M_{PW}(g,d(g))
\end{eqnarray}
where $d(g)$ is the domain of level $g$ (as defined in \eqref{domain}). It can easily be seen that the defined $M_v \in [0,1]$. Fig. \ref{fig_simPW} to show MV using this method.\\

\subsubsection{Fuzzy Membership Value- Center-of-weights Method}

\begin{figure}[H]
\centering
\subfloat[]{\includegraphics[scale = .45]{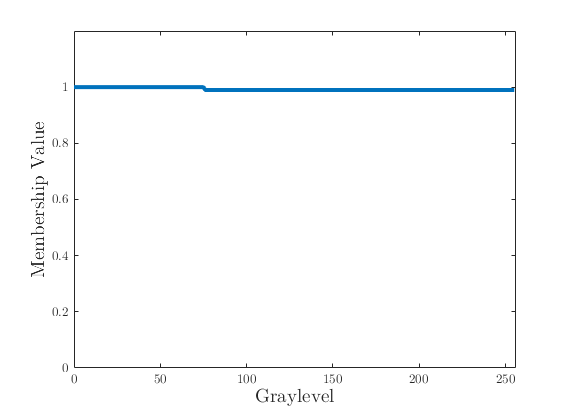}
\label{CW Upper MV}}
\hfil
\subfloat[]{\includegraphics[scale = .45]{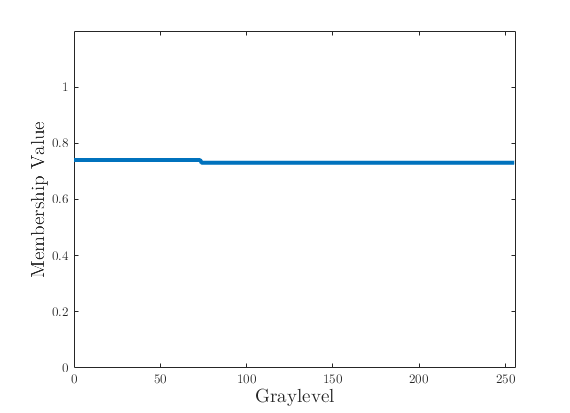}
\label{CW Lower MV}}
\caption{Memebership Value using center-of-weights method : (a) upper MV and (b) lower MV}
\label{fig_simCW}
\end{figure}

The weighted mean (also called as the center-of-weight) of a given set of values represents the point at which the entire weighted set of values may be assumed to be concentrated\cite{p2}. Using this, Membership Value may be defined as 
\begin{eqnarray}\label{cowmv1}
M_v(g) = M_{CW}(d(g))
\end{eqnarray}
where $d(g)$ is the domain of gray level $g$ and the function $M_{CW}(i)$ is defined as \begin{eqnarray}\label{cowmv2}
M_{CW}(i) = F_i(\overline{g_i})
\end{eqnarray}
where 
\begin{eqnarray}\label{cowmv3}
\overline{g_i} = \frac{\sum\limits_{g = 0}^{L-1}min\{F_i(g),H(g)\}\cdot g}{\sum\limits_{g = 0}^{L-1}min\{F_i(g),H(g)\}}
\end{eqnarray}
Equation \eqref{cowmv3} defines $\overline{g_i}$ as the center-of-weight of the area of overlap of the histogram and the component function $F_i$ of the upper or lower MF, and the function $M_{CW}(i)$ computes the membership value of $\overline{g_i}$ in \eqref{cowmv2}. Since the component membership function $F_i$ is an symmetric Gaussian, the value of $M_{CW}(i)$ decreases as the distance between $\overline{g_i}$ and the center of the symmetric Gaussian  $\mu_i$ increases. Eqautions \eqref{cowmv1} and \eqref{cowmv2} together imply that the membership value of all gray levels in the reach of $F_i$ is given by $M_{CW}(i)$. Hence, localized equalization of the input histogram may be performed according to an adjusted value (the fuzzy MV). This is because uniform probability is assigned to all gray levels in the reach interval by the PDF generation method. Fig. \ref{fig_simCW} shows the obtained MV using this method.\\

\subsubsection{Fuzzy Membership Value- Area Method}

The area method is based on extent of overlap of a particular gray levels domain and the normalized histogram values. Membership Value of a gray level $g$ is obtained by using-\begin{eqnarray}\label{areamv1}
M_v(g) = M_A(d(g))
\end{eqnarray} 
where $d(g)$ is the domain of the level $g$ and the function $M_A(i)$ is defined as \begin{eqnarray}\label{areamv2}
M_A(i) = \frac{\sum\limits_{g = 0}^{L-1}min\{F_i(g),H(g)\}}{\sum\limits_{g = 0}^{L-1}F_i(g)}
\end{eqnarray}
Equation \eqref{areamv2} shows that as the area of overlap between the minimum of functions - $F_i(g)$ and $H(g)$ and fitted function increases, the value of fuzzy membership function increases. The nature of increase depends on the relative magnitude of overlap and can be use for controlling the PDF formation.\par
\begin{figure}[H]
\centering
\subfloat[]{\includegraphics[scale = .45]{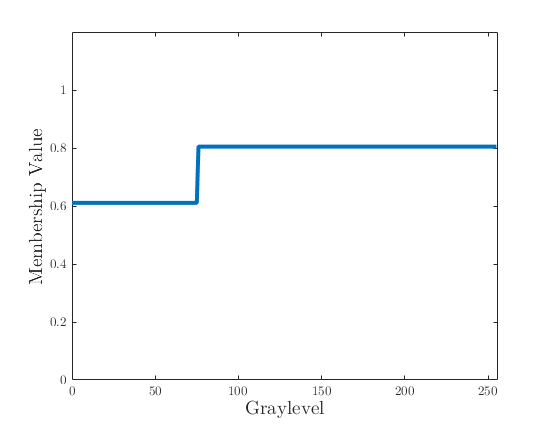}
\label{Ar Upper MV}}
\hfil
\subfloat[]{\includegraphics[scale = .45]{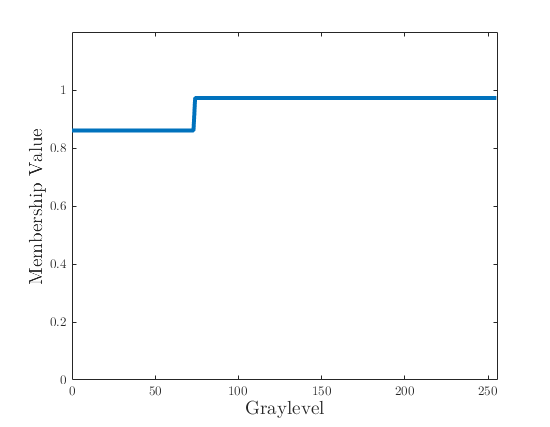}
\label{Ar Lower MV}}
\caption{Memebership Value using area method : (a) upper MV and (b) lower MV}
\label{fig_simAr}
\end{figure}

If the best-fit gaussian function is not a good approximation of the histogram in the neighbourhood of gray level $g$, then the overall magnitude of membership function is small. However, if the domain of the function sufficiently approximates the histogram, the obtained membership values may tend to be large. Fig. \ref{fig_simAr} shows the upper and lower MV obtained from this method.\\

\subsubsection{Fuzzy Membership Value- KM Algorithm}

We aim to divide the given histogram in $n$ appropriate divisions, which we call as \textit{clusters}. We find the centroid of these clusters $v_j$ $(j\in\{1,\cdots,n\} )$ using the KM algorithm, followed by type reduction from type-2 to type-1 which is described in this section:\cite{p3}\par
\begin{figure}[H]
\centering
\subfloat[]{\includegraphics[scale = .6]{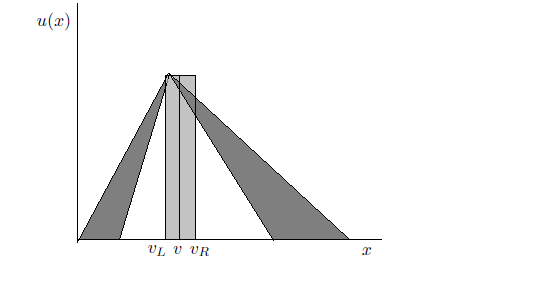}
\label{Center of IT2 fuzzy set}}
\hfil
\subfloat[]{\includegraphics[scale = .45]{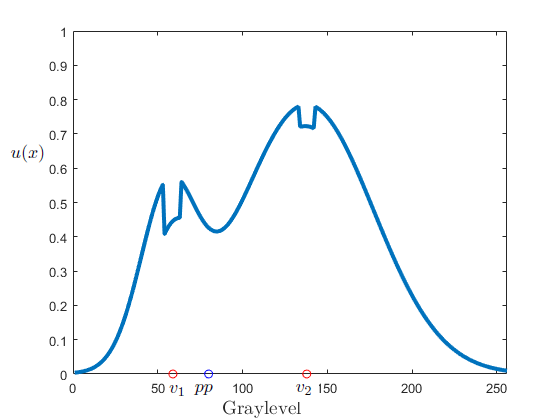}
\label{MV_KM}}
\caption{Illustration of Karnik-Mendel method : (a) center of IT2 fuzzy set, (b) type-reduction}
\label{fig_simKarMen}
\end{figure}

To get $v_j$, as shown in Fig. \ref{fig_simKarMen}\subref{Center of IT2 fuzzy set}
Karnik \textit{et al.} have suggested an iterative algorithm which estimates both the ends of an interval, that is, estimating $v_R$ and $v_L$ for each of the desired centroid $v_j$. It can be estimated by arranging the pattern set in ascending order and finding the location between upper and lower membership for a pattern set.  It is required for an ascending ordered pattern set to represent a left value $v_L$ and a right value $v_R$ of an interval cluster center. We apply the iterative algorithm to estimate the cluster center. It first estimates $v_R$ and $v_L$ for every centroid, from which $v_j$ can be calculated easily.\cite{p3}

 \par

\begin{algorithm}
\caption{Finding maximum of center $v_j$ : $v_R$}\label{algo}
\begin{algorithmic}[1]

\State Set fuzzfier $m$ = arbitrary. (We chose 2).
\State Compute centroid $v_j$ using \eqref{u(x_i)} and \eqref{compute centroid}
\begin{eqnarray}\label{u(x_i)}
u(x_i) = \frac{umf(x_i)+lmf(x_i)}{2}
\end{eqnarray}
\begin{eqnarray}\label{compute centroid}
v_j = \frac{\sum\limits_{i = 1}^N x_i\cdot {u(x_i)^m}}{\sum\limits_{i = 1}^N {u(x_i)^m}}
\end{eqnarray}
\State Sort pattern indices for all N patterns $(i=1,\cdots,N)$ in ascending order. Sorted feature = $x_1 \leq \cdots \leq x_N$

\State Set comparison = false;
\While {comparison = false}
	\State Find interval index $k\,(1\leq k\leq N-1)$ such that $x_k\leq v_j\leq x_{(k+1)}$; 
    \ForAll {patterns} 
	  \If {$(i\leq k)$} 
		\State Set primary membership $u_j(x_i)\,=\,lmf(x_i)$;
	  \Else
		\State Set primary membership $u_j(x_i)\,=\,umf(x_i)$;
	  \EndIf
	\EndFor
\State Compute maximum of center candidate $v_j'$ using the modified $u_j(x_i)$ obtained above;
	  \If {$(v_j = v_j')$}
		\State Set Comparison = TRUE;
	  \Else
		\State Set $v_j\,=\,v_j'$;
	  \EndIf 
\EndWhile
\State \textbf{end}
\State Set $v_R\,=\,v_j'$
\end{algorithmic}
\end{algorithm}

\begin{algorithm}
\caption{Finding maximum of center $v_j$ : $v_L$}\label{algo2}
\begin{algorithmic}[1]
	  \If {$(i\leq k)$} 
		\State Set primary membership $u_j(x_i)\,=\,umf(x_i)$;
	  \Else
		\State Set primary membership $u_j(x_i)\,=\,lmf(x_i)$;
	  \EndIf
	
\end{algorithmic}
\end{algorithm}
\begin{figure}[H]
\centering
{\includegraphics[scale = .45]{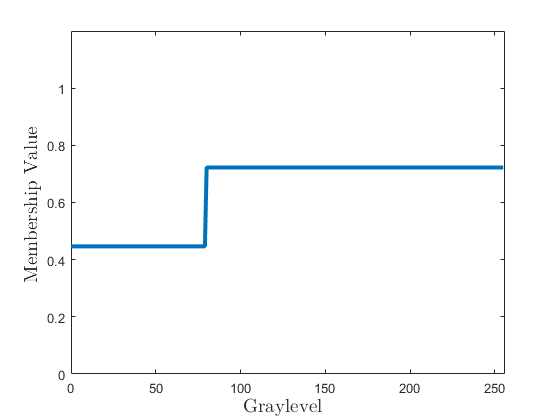}
\label{MV KM}}
\caption{Membership Value using Karnik-Mendel method}
\label{fig_simKMN}
\end{figure} 
 \par
 A crisp center for estimated center $v_j$ is simple to compute as we are working on IT2 fuzzy sets.
 \begin{eqnarray}
 v_j = \frac{v_L + v_R}{2}
 \end{eqnarray}
Fig. \ref{fig_simKarMen}\subref{Center of IT2 fuzzy set} shows the center $v_j = v$ for one cluster. \par
 In our proposed algorithm, we perform type reduction in order to estimate cluster centers. For this, left memberships ${u_j}^L(x_i)$ and right memberships ${u_j}^R(x_i)$ for all patterns have already been estimated to organize $v_L$ and $v_R$, respectively. Therefore, type reduction can be achieved using ${u_j}^L(x_i)$ and ${u_j}^R(x_i)$ (\eqref{type reduction KM} and \eqref{TypeRed KM}) and hence membership value of a gray level $g$ can be found out using \eqref{MV KM}. Note that $C$ is the total number of clusters.\cite{p3}
 \begin{eqnarray} \label{type reduction KM}
 u_j(x_i) = \frac{{u_j}^L(x_i) + {u_j}^R(x_i)}{2}, \qquad \qquad j\,=\,1,\cdots, C
 \end{eqnarray}
 \begin{eqnarray}\label{TypeRed KM}
 T(g) = \sum \limits_{j=1}^C u_j(g) ,\qquad \qquad g \in \, cluster \, j
 \end{eqnarray}
 \begin{eqnarray}\label{MV KM}
 M_v(g) = T(v_j), \qquad\qquad \, \forall \, g \in cluster \, j
 \end{eqnarray}

\subsection{PDF Generation}

Gray levels which have high frequency of occurrence concentrate near the peak(s) of the histogram of the image. Thus, it is difficult to differentiate between those levels. This problem can be solved by using the formula for the desired PDF which assigns higher probability to those gray levels which are away from the peak(s) of membership function. Probability value assigned is directly proportional to the distance of gray level from the peak(s) of histogram.\par
\textbf{PDF for Point-wise, Center of weights and Area methods:}\par
We propose the following linear equation for generation of a suitable PDF for histogram specification using the fuzzy membership values (obtained through methods as described in the previous section). The un-normalized probability of occurrence ${P_D}^U(g)$ for the upper MF of a given gray level $g$ in the output histogram may be calculated as  
\begin{eqnarray} \label{upperPDF}
{P_D}^U(g) = \begin{cases}
            T + 2{M_v}^U(g)\left[\left(\frac{\mu_{d(g)}+c_{i_1}}{2}\right)-g\right]  & \text{if } g < \mu_{d(g)} \\ \\
            T - 2{M_v}^U(g)\left[\left(\frac{\mu_{d(g)}+c_{i_2}}{2}\right)-g\right]  & \text{if } g \geq \mu_{d(g)}  
       \end{cases} 
\end{eqnarray}
where $T$ is the largest gray level in the (normalized) image, i.e, $L-1$, ${M_v}^U(g)$ is the fuzzy membership value of $g$ calculated for the upper MF, ${\mu}_i$ is the center of the $i^{th}$ component membership function $F_i$, $d(g)$ is the domain function of gray level $g$, $c_{i_1}$ and $c_{i_2}$ are the crossover points for $d(g)$.\par 
In an analogous manner, the un-normalized probability of occurrence ${P_D}^L(g)$ for the lower MF may be obtained as  
\begin{eqnarray}\label{lowerPDF}
{P_D}^L(g) = \begin{cases}
            T + 2{M_v}^L(g)\left[\left(\frac{\mu_{d(g)}+c_{i_1}}{2}\right)-g\right]  & \text{if } g < \mu_{d(g)} \\ \\
            T - 2{M_v}^L(g)\left[\left(\frac{\mu_{d(g)}+c_{i_2}}{2}\right)-g\right]  & \text{if } g \geq \mu_{d(g)}  
       \end{cases} 
\end{eqnarray}
where ${M_v}^L(g)$ is the fuzzy membership value of $g$ calculated for the lower MF and all other variables has same meaning as before. Due to interval type-2 nature of MF and the linear nature of the MV computation techniques, the lower and upper PDFs may be de-fuzzified without centroid computation through simple mean computation as follows:
\begin{eqnarray}\label{PDFfinal}
P_D(g) = \frac{1}{2}\left({P_D}^U(g)+{P_D}^L(g)\right)
\end{eqnarray} 
where $P_D(g)$ is the final un-normalized PDF. We note that in this case, the PDF itself represents a type-1 fuzzification of the gray level probabilities. The usage of crisp outputs is inherited in the HS algorithm, which samples the PDF according to discrete gray levels. \par
Geometrically, \eqref{upperPDF} and \eqref{lowerPDF}, each consider the distance between the gray level, and the mean of the domain functions' symmetric Gaussian peak and the crossover point, depending on which arm of the symmetric Gaussian the gray level lies. This value is then scaled by the fuzzy membership value and shifted by an amount equal to the largest gray level in the image, Finally, we note that the distribution obtained using \eqref{PDFfinal} may be normalized to obtain a PDF $\widetilde{P_D}(g)$ suitable for histogram specification. \par

\begin{figure}[H]
\centering
\subfloat[]{\includegraphics[scale = .45]{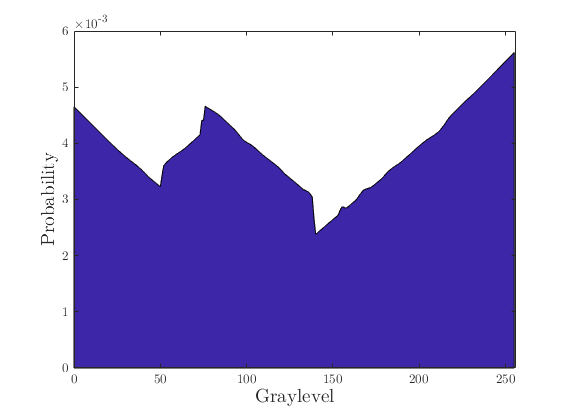}
\label{PDF PW}}
\hfil
\subfloat[]{\includegraphics[scale = .45]{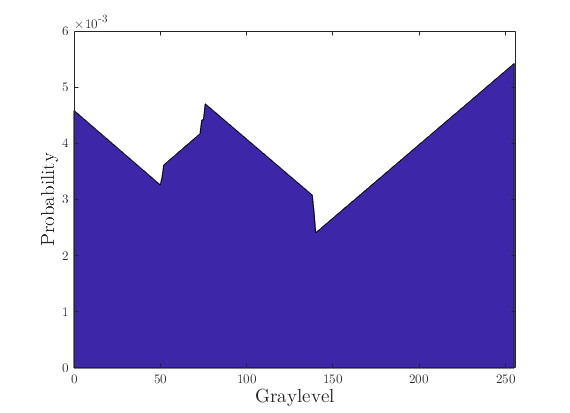}
\label{PDF COW}}
\hfil
\subfloat[]{\includegraphics[scale = .45]{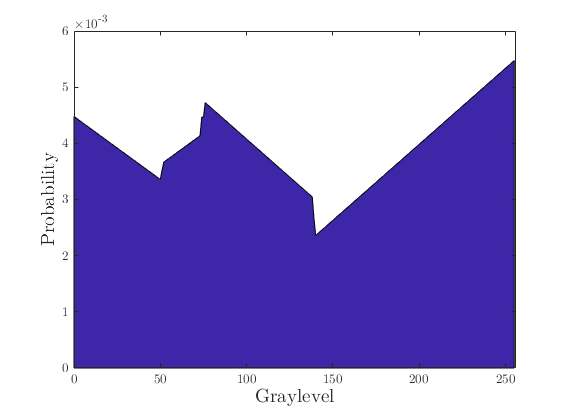}
\label{PDF Ar}}
\hfil
\subfloat[]{\includegraphics[scale = .45]{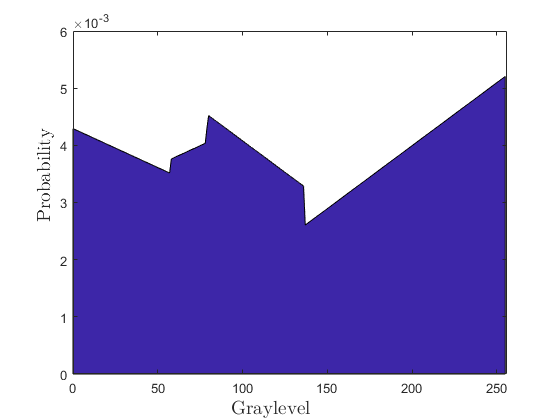}
\label{PDF KM}}
\caption{Illustration of the suitable PDF generated for HS using : (a) point-wise, (b) center-of-weights, (c) area, and (d) KM method}
\label{fig_sim_ARKMPDF}
\end{figure}

\newpage
\textbf{PDF for the Karnik-Mendel method:}\par It can be found out using the similar concepts used for the other 3. The following formula can  be used to find the appropriate PDF 
\begin{eqnarray}\label{KMPDF}
{P_D}(g) = \begin{cases}
            T + 2{M_v}(g)\left[\left(\frac{v_j+start_j}{2}\right)-g\right]  & \text{if } g < v_j \\ \\
            T - 2{M_v}(g)\left[\left(\frac{v_j+end_j}{2}\right)-g\right]  & \text{if } g \geq v_j  
       \end{cases} 
\end{eqnarray}
where $v_j$ is the center of cluster $j$, $start_j$ is the starting gray value $g$ for the cluster $j$ and $end_j$ is the last gray value $g$ for the cluster $j$ which can easily be found using the partition point(s) obtained earlier.\\

\subsubsection*{Effect of Fuzzy MV Method on PDF Generation}
It is clear from \eqref{upperPDF}, \eqref{lowerPDF} and \eqref{PDFfinal} that PDF depends crucially on the fuzzy membership values and hence, on the value generation method. The proposed equations are linear in nature and it provides less value of probability for those gray levels which have distribution of significant height of the histogram and more value to others. This nature is justified as our aim is to enhance that portion of image which has no intensity corresponding to the particular gray levels. By providing higher value of PDF for such gray levels, we try to make image clearer by using histogram specification. Since the choice of value computation method is left to the application, this provides flexibility in generation of the output PDF and allows for finer control over the distribution. For example, the \textit{point-wise} method assigns higher probabilities to less frequently occurring gray levels leading to better distribution in the output histogram, i.e. if the difference is small, then the membership value can be assigned a large value so that histogram spreads over the gray levels, more effectively. The \textit{center-of-weights} method manipulates the closely spaced gray levels, thereby performing localized histogram equalization.The \textit{area} method assigns equal probability to closely-spaced gray levels and also controls the degree of spreading of gray levels between closely-spaced regions. In general, this method has smaller variation in gray level spreading. Fig. \ref{fig_sim_ARKMPDF} shows different PDF obtained from the proposed 4 methods.

\subsection{Histogram Specification}

\begin{figure}[H]
\centering
\subfloat[]{\includegraphics[scale = .38]{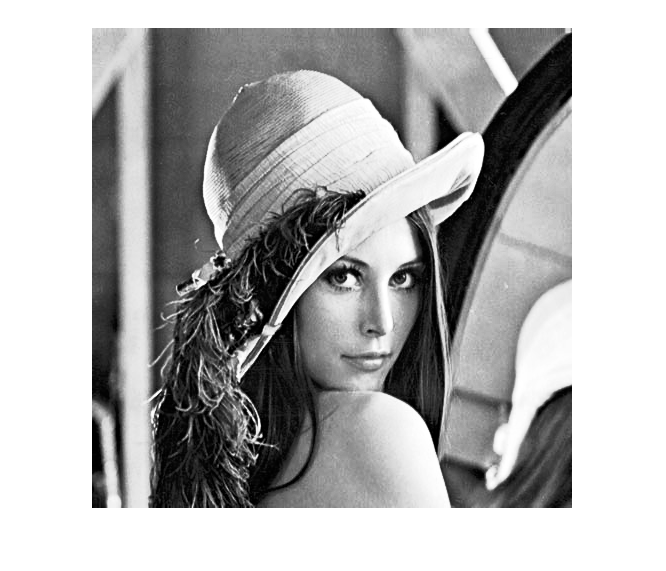}
\label{EM PW}}
\hfil
\subfloat[]{\includegraphics[scale = .38]{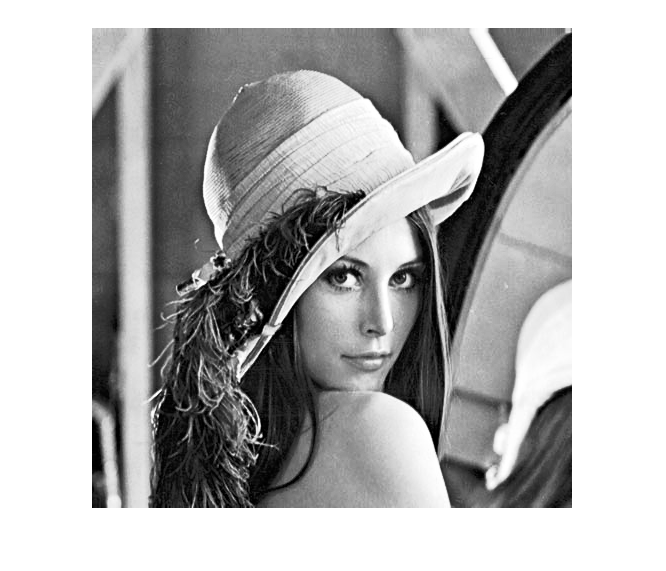}
\label{EM COW}}
\label{fig_sim PWCOW}
\hfil
\\ \subfloat[]{\includegraphics[scale = .38]{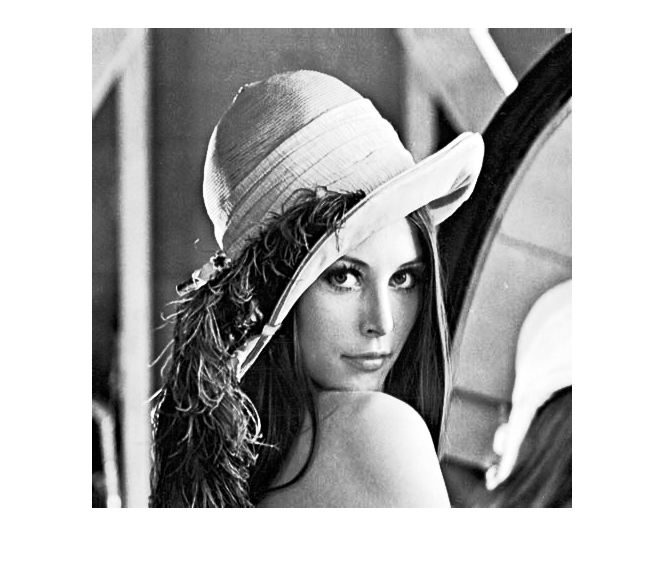}
\label{EM AR}}
\hfil
\subfloat[]{\includegraphics[scale = .30]{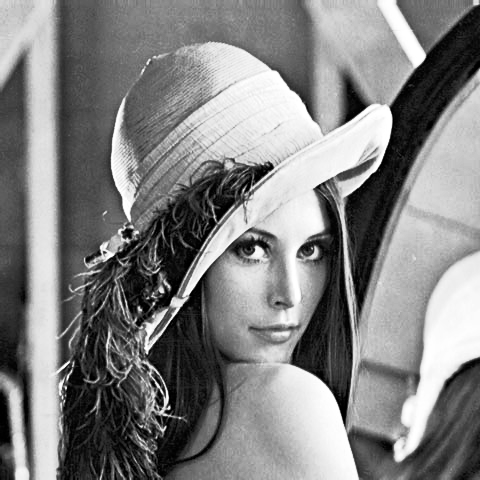}
\label{EM KM}}
\caption{Enhanced Images after Histogram Specification : (a) point-wise, (b) center-of-weights, (c) area, and (d) KM method}
\label{fig_sim_EMARKM}
\end{figure}

This is the final stage in our proposed algorithm. Usual Histogram Specification is done using the PDF generated in previous steps. Fig. \ref{fig_sim_EMARKM} shows the final results obtained from the proposed 4 methods. We note that the algorithm does not require any external inputs in addition to the input image during its execution. Therefore, the proposed algorithm may find suitable fuzzy MFs for a given image automatically and without need of prior knowledge about the number of MFs. Consequently, the algorithm may find the proper desired PDF for each given image and obtain the contrast enhanced image automatically. Thus, the proposed algorithm can find suitable fuzzy membership function for a given image automatically and without need of prior knowledge of the number of fuzzy membership function. Consequently, the algorithm can find the proper desired PDF for each given image and obtain the contrast enhanced image automatically.

\subsection{Summary}

\begin{tabular}{*5l}    
\bottomrule
\hline
Stage 1   & Smoothen and normalize the histogram of the input image. \\
&  Find the appropriate partition to divide the histogram. \\
&  Fit the symmetric Gaussian(s). \\
\hline 
Stage 2	 & -Generate UMF and LMF and hence obtain FOU.\\ 
\hline 
Stage 3  & Generate Membership Values from FOU using either of 4 methods- \\
& - Point-wise method \\
& - Center-of-weights method \\
& - Area method \\
& - KM method \\
\hline
Stage 4 & Generate the Probability Distribution Function (PDF) using the provided formulae. \\
\hline
Stage 5 & Apply Histogram Specification using the generated PDF. \\
\\ \bottomrule
 \hline
\end{tabular}\\ \par
In the described methodology, in stage-1, we are smoothing, normalising and fitting the appropriate number of Gaussian(s) to divide the image pixel into clusters. For instance, pixels can be clustered into dark and bright. As the fitting is done over a normalised histogram, it can be considered as calculating membership function corresponding to each gray level. Later, UMF and LMF was found to implement the interval type-2 fuzzy to obtain the membership value and probability density function. PDF equation is designed in such a way that it gives high probability to the gray levels having less number of pixels and low probability to the gray levels having high number of pixels. This is how, it enhances the image more accurately than the histogram equalisation, which provides equal probability to all the gray levels. 

\section{Experiment Results and Discussions}
In this section, we demonstrate the results obtained using the proposed algorithm. 	In addition to it, we compared the obtained results with 3 already present methods namely, Histogram Equalisation (HE), Recursive
Mean-Separate Histogram Equalization (RMSHE) and Brightness Preserving Fuzzy Histogram Equalization (BPFHE). Quantitative analysis is done using the image quality index, called \textit{Average Information Content} (AIC). The information content is related to the number of binary decisions required to find the information. The number of binary decisions (number of questions whose answer is yes/no) required to find the correct element in a set of $N$ elements is:
\begin{eqnarray}
n_q = log_2N = -log_2p
\end{eqnarray}
In general, the elements are not equally likely; they have different probabilities, $p_i$. Tribus(1961) then generalises the formula here above by introducing the concept of "\textit{surprisal $h_i$}". \begin{eqnarray}
h_i = -log_2p_i
\end{eqnarray}
On this basis, Shannon introduced the \textit{uncertainty measure} (also called \textit{entropy}, which is the average of all surprisals $h_i$ weighted by their occurrence $p_i$. \begin{eqnarray}
H = \sum\limits_i p_ih_i = -\sum\limits_i p_ilog_2p_i
\end{eqnarray}
Quantitative comparison of images was performed on the basis of the average information content (AIC) measure. For images, it can be written more precisely as: \begin{eqnarray}
AIC = -\sum\limits_{k = 0}^{L - 1} p(g_k)log_2(p(g_k))
\end{eqnarray}
where $p(g_k)$ is the PDF value for the $k^{th}$ gray level. In general, AIC increases with an increase in the information content of the image. In other words, a higher score indicates a \textit{richer}, more detailed image. Table 1 shows the AIC values of different methods already present along with our proposed methods\cite{r1}. Note that the values written in bold represents the maximum AIC value among all of described methods. \par
\begin{figure}
\centering
\includegraphics[scale=.35]{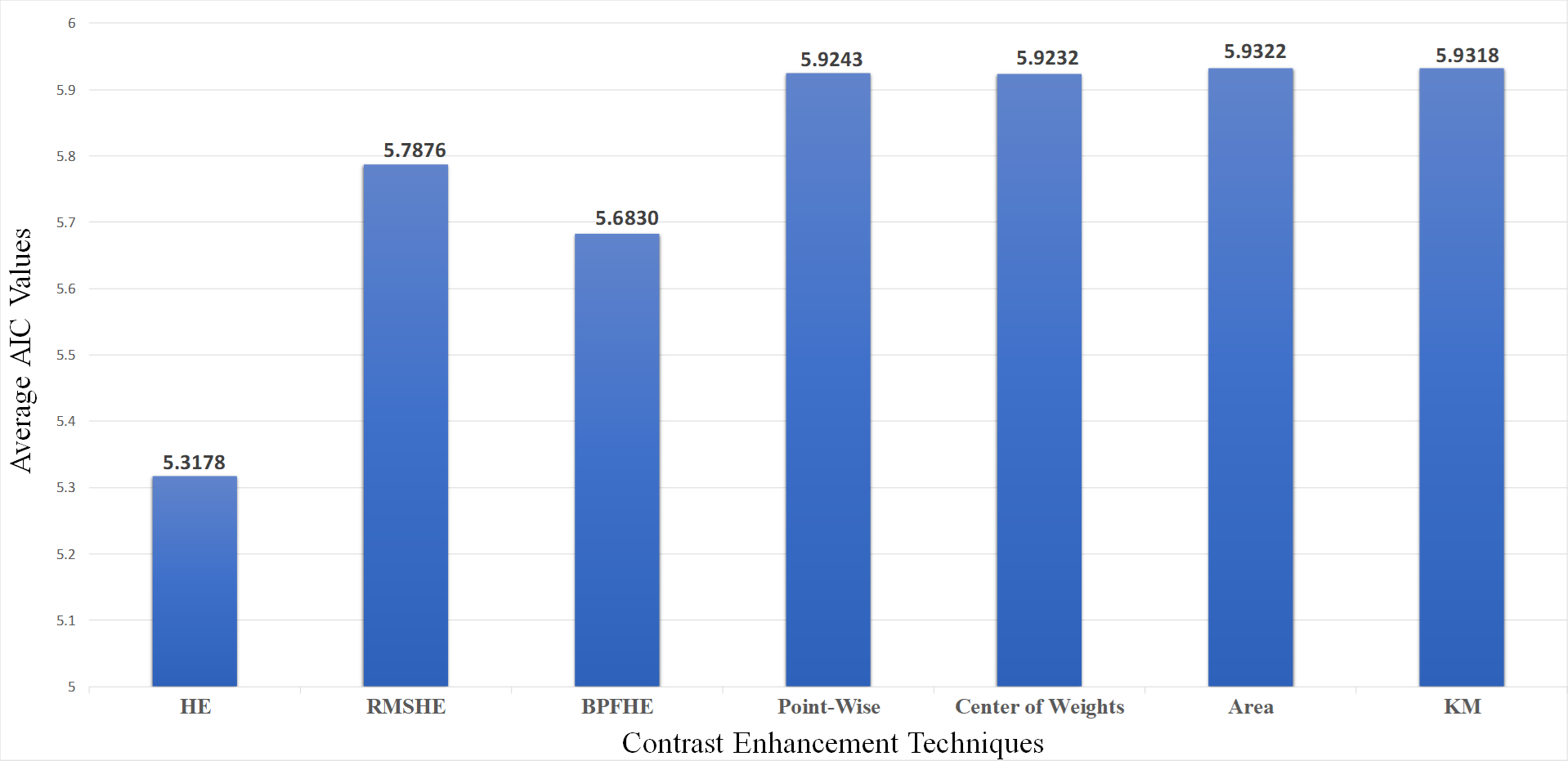}
\caption{{Comparision of AIC value for all images}}\label{compAIC}
\end{figure}

 
A bar graph shown in Fig. \ref{compAIC} depicts the comparison of AIC values of all methods.\par
\begin{figure}[H]
\centering
\subfloat[]{\includegraphics[scale = .72]{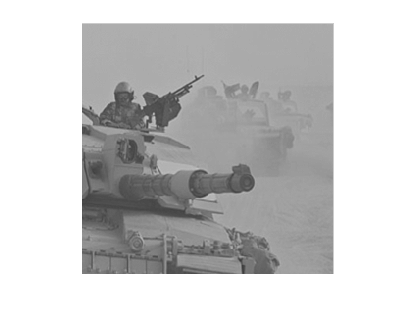}
\label{Original Tank}}
\subfloat[]{\includegraphics[scale = .6]{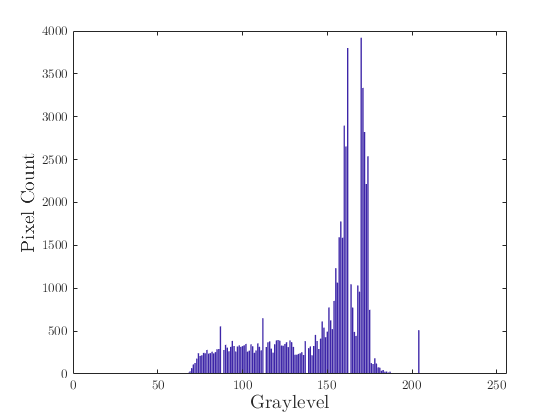}
\label{Histogram tank}}
\caption{\lq{Tank}\rq image : (a) original image, (b) input histogram}
\label{tank orig and hist}
\end{figure}
\begin{figure}[H]
\centering
\subfloat[]{\includegraphics[scale = .58]{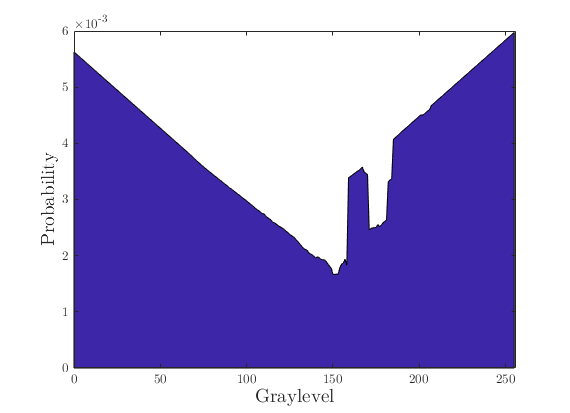}
\label{COW Tank}}
\subfloat[]{\includegraphics[scale = .58]{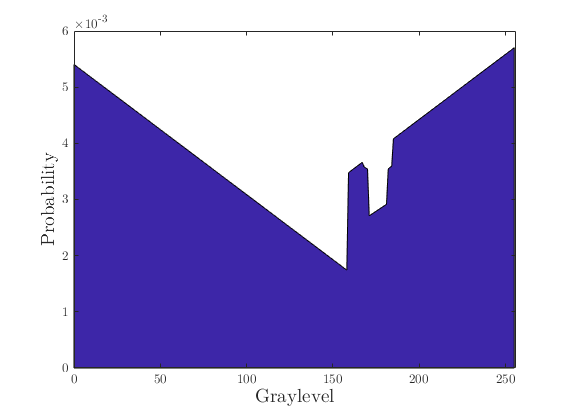}
\label{COW tank}}
\hfil
\subfloat[]{\includegraphics[scale = .58]{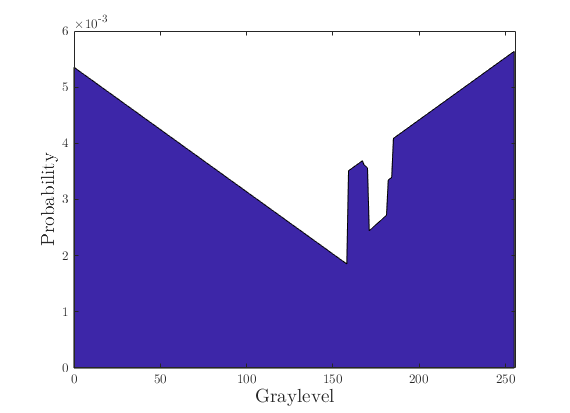}
\label{ar tank}}
\subfloat[]{\includegraphics[scale = .58]{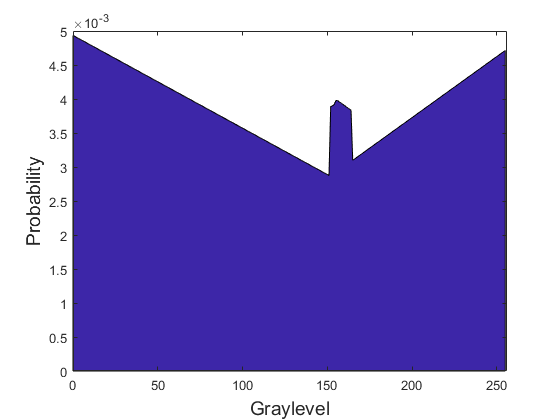}
\label{km tank}}
\caption{Generated PDF for \lq{Tank}\rq image using : (a) point-wise, (b) center-of-weights, (c) area, and (d) KM method}
\label{fig_simtank2}
\end{figure}

We present the analysis of our algorithm and compare it with already present 3 methods on 3 images - \lq Tank\rq , \lq Classroom\rq , \lq Vendovka\rq . As seen in Fig.\ref{tank orig and hist}\subref{Histogram tank}, histogram of image \lq Tank\rq  is concentrated in the middle (near gray level $150$). As desired, we got the PDFs (Fig. \ref{fig_simtank2}) which has the least value around $150^{th}$ gray level. This spreads the final histogram of image to left as well as right of $150^{th}$ gray level, enhancing the contrast of the original image. 
AIC value is found to be increased by $13\%$ as compared to that of HE. KM method worked best for this image.\par
\begin{figure}[H]
\centering
\subfloat[]{\includegraphics[scale = .72]{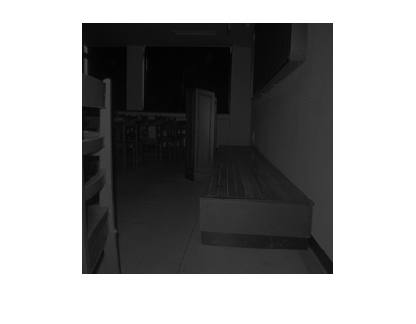}
\label{Original Class}}
\hfil
\subfloat[]{\includegraphics[scale = .6]{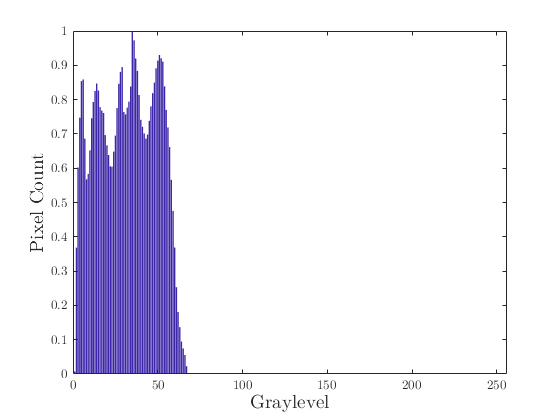}
\label{Histogram Class}}
\caption{\lq{Classroom}\rq image : (a) original image, (b) input histogram}
\label{fig_simclass1}
\end{figure}
\begin{figure}[H]
\centering
\subfloat[]{\includegraphics[scale = .59]{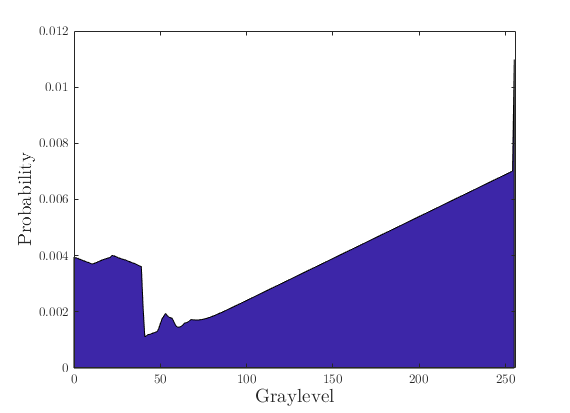}
\label{COW class}}
\subfloat[]{\includegraphics[scale = .59]{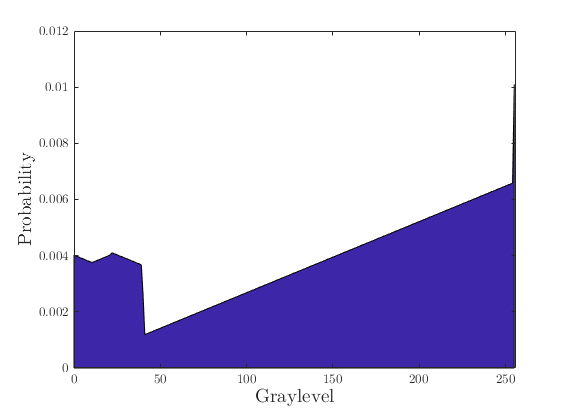}
\label{COW class}}
\hfil
\subfloat[]{\includegraphics[scale = .59]{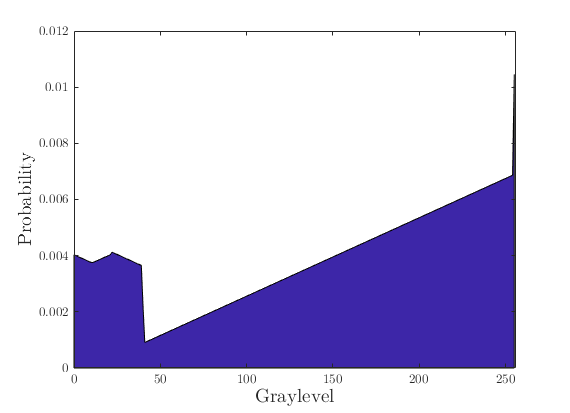}
\label{ar class}}
\subfloat[]{\includegraphics[scale = .59]{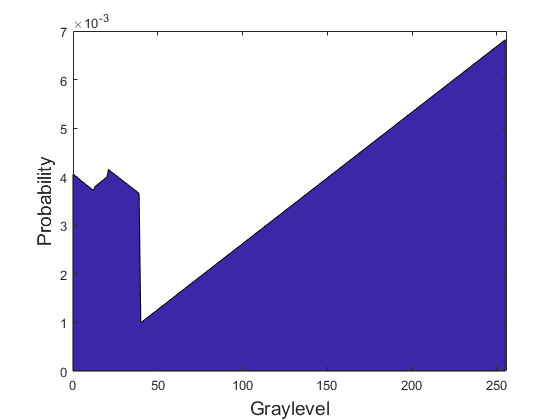}
\label{km class}}
\caption{Generated PDF for \lq{Classroom}\rq image using : (a) point-wise, (b) center-of-weights, (c) area, and (d) KM method}
\label{fig_simclass3}
\end{figure}

For image \lq Classroom\rq , histogram of original image is found to be concentrated before around $70^{th}$ gray level (Fig.\ref{fig_simclass1}\subref{Histogram Class}). The PDFs obtained from our proposed algorithm has it's least value around $70^{th}$ gray level and it increases as gray level increased from $70$ to $255$ (Fig.\ref{fig_simclass3}). It spreads the histogram to the right side, hence enhancing the contrast of original image. AIC value is increased by $1.07\%$ as compared to that of HE. Interestingly, KM method worked best for this image too.\par
\begin{figure}[H]
\centering
\subfloat[]{\includegraphics[scale = .42]{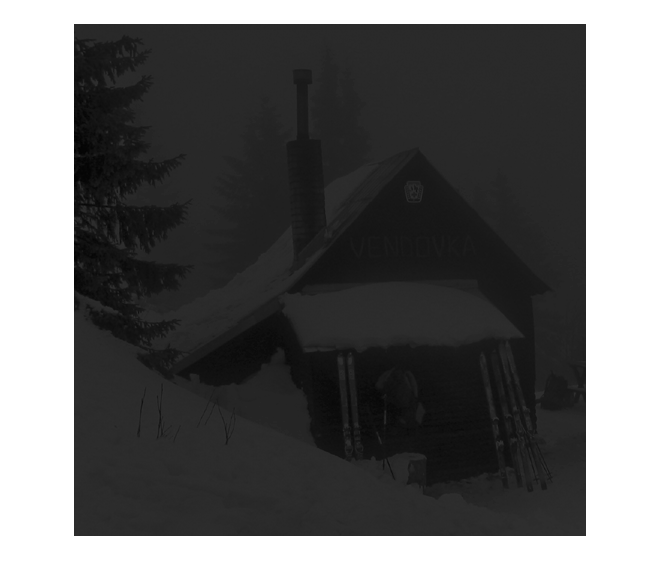}
\label{Original vendovka}}
\hfil
\subfloat[]{\includegraphics[scale = .45]{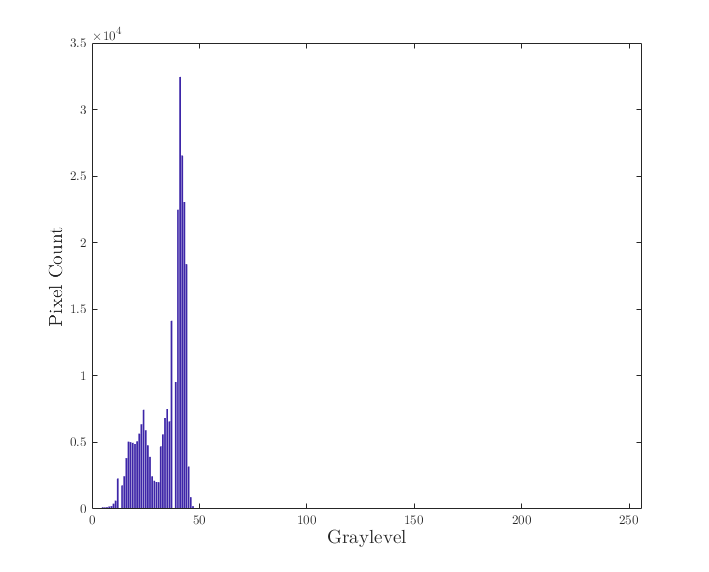}
\label{Histogram vendovka}}
\caption{\lq{Vendovka}\rq image : (a) original image, (b) input histogram}
\label{fig_simven1}
\end{figure}
\begin{figure}[H]
\centering
\subfloat[]{\includegraphics[scale = .59]{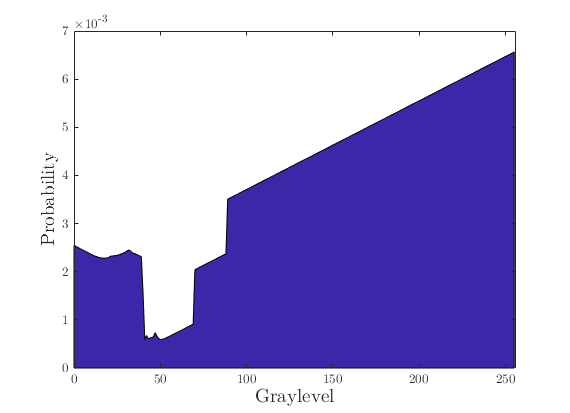}
\label{COW ven}}
\subfloat[]{\includegraphics[scale = .59]{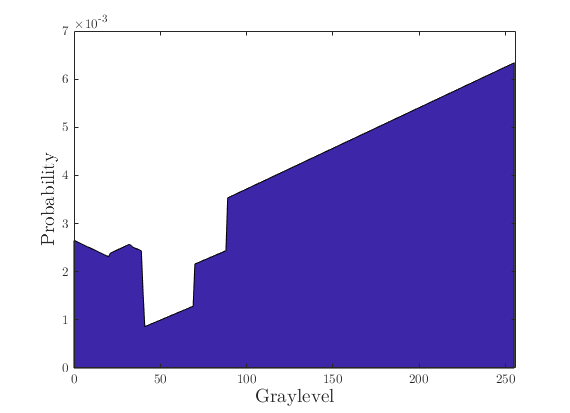}
\label{COW ven}}
\hfil
\subfloat[]{\includegraphics[scale = .59]{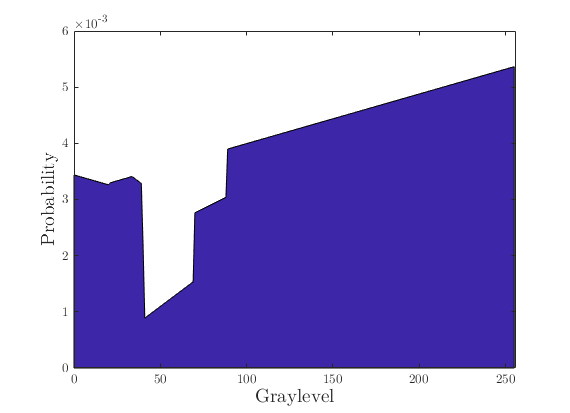}
\label{ar ven}}
\subfloat[]{\includegraphics[scale = .59]{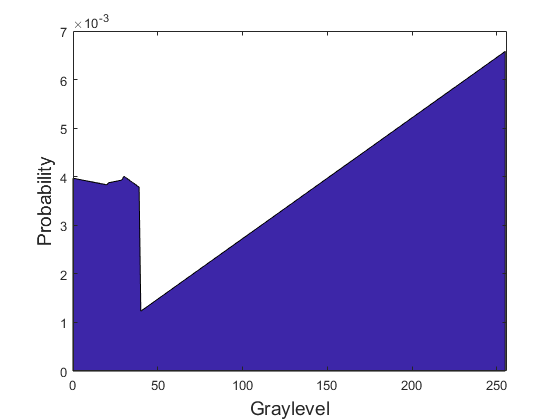}
\label{km ven}}
\caption{Generated PDF for \lq{Vendovka}\rq  image using : (a) point-wise, (b) center-of-weights, (c) area, and (d) KM method}
\label{fig_simven3}
\end{figure}

The histogram of image \lq Vendovka\rq is concentrated mostly around $35^{th}$ gray level (Fig.\ref{fig_simven1}\subref{Histogram vendovka}). The PDFs obtained dips near $40^{th}$ gray level, and increases as value of gray level increases (Fig.\ref{fig_simven3}). This spreads the histogram of final image to the right, enhancing the image. AIC value is increased by $1.62\%$ in comparison to that of HE. However, contrary to earlier cases, this image shows the best AIC value after being processed by Point-wise and Center of weights methods.

		\par

\vspace*{1 cm}
\begin{table}[H]
\renewcommand{\arraystretch}{1.3}
\caption{Experimental AIC Values} 
\label{aic_value}
\centering
\begin{tabular}{ M{2 cm}M{1.5cm} M{1.5cm} M{1.5cm} M{1.5cm} M{1.5cm} M{1.5cm} M{1.5cm} }
\hline \hline

{Image Name} & {HE} &{RMSHE}&{BPFHE} & {Point-Wise} & {\centering Center of Weights} &{Area} & {KM Algo.} \\
 \hline
 Lena   &\small 5.9771    &\small 7.0824 &\small 7.2748 & \small 7.3034 &\small 7.3020 &\small \textbf{7.3035} & \small {7.3015}\\
 {Classroom}   &\small 5.7884 &\small 5.6951 &\small 5.3573 &\small 5.8476 &\small {5.8476} &5.8476 &\small \textbf{5.8490}\\
{Wood} &\small 4.6507&\small 4.7357&\small 4.2906&\small 4.8248 &\small 4.8279&\small 4.8267&\small\textbf{4.8290}\\
{Vendovka}    &\small 4.4804&\small 4.4704&\small 4.1791&\small\textbf{4.5527}&\small \textbf{4.5527}&\small{4.5485}&\small 4.5480\\
{Tank}        &\small 5.3293&\small	5.8783&\small	5.8884&\small	6.0098&\small	6.0018&\small	6.0098&\small	\textbf{6.0222}\\
 {Park}	&\small 5.6881	&\small 5.8592	&\small 5.6391&\small	6.0245&\small	\textit{6.0345}&\small	6.0355&\small	\textbf{6.0440}\\
{Hilly Area}	&\small 4.6946	& \small 5.6468&\small	5.6507	&\small 5.6537	&\small{5.6397}	&\small \textbf{5.6778}	&\small 5.6499\\
{House}	&\small 4.8299	&\small 4.9056	&\small 4.5675	&\small 4.9811	&\small 4.9904	&\small 4.9928	&\small\textbf{5.0093}\\
{Children}	&\small 5.9743	&\small 7.0063	&\small 7.2546	&\small 7.3014	&\small 7.3039	&\small 7.3048	&\small\textbf{7.3070}\\
{Nature}	&\small 5.7658&\small	6.5970	&\small 6.7286&\small	6.7435	&\small 6.7314	&\small \textbf{6.7756}	&\small{6.7575}\\
 \hline
\end{tabular} \par 

\end{table}
\clearpage

%\begin{center}
%\textbf{Tank: Image Processed using different methods}
%\end{center}

\begin{figure}[H]
\centering
\subfloat[]{\includegraphics[scale = .72]{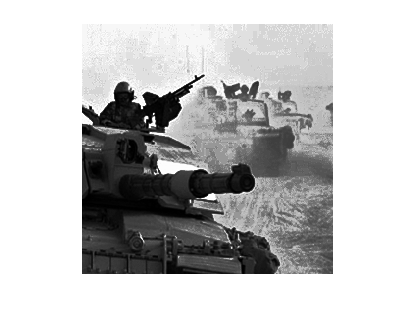}
\label{PW Tank}}
\subfloat[]{\includegraphics[scale = .72]{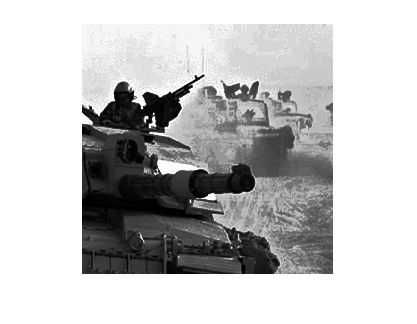}
\label{COW tank}}
\hfil
\subfloat[]{\includegraphics[scale = .72]{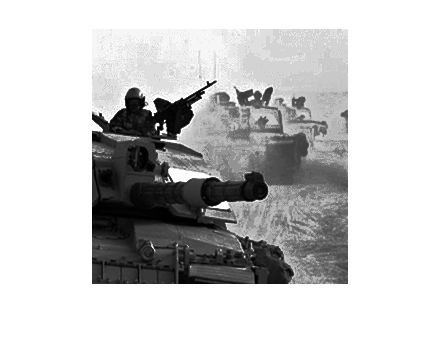}
\label{ar tank}}
\subfloat[]{\includegraphics[scale = .72]{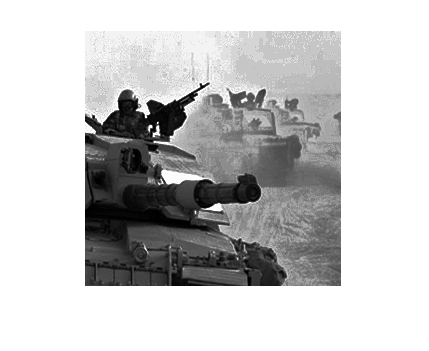}
\label{km tank}}
\caption{Enhanced image of \lq{Tank}\rq for : (a) point-wise, (b) center-of-weights, (c) area, and (d) KM method}
\label{en_image}
\end{figure}

\newpage
\vspace*{2 cm}

%%%%%%%%%%%%
\newpage

\begin{figure}[H]
\centering
\subfloat[]{\includegraphics[scale = .72]{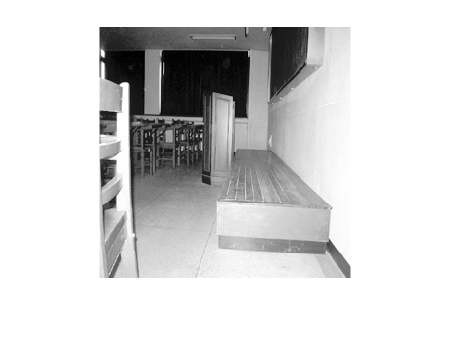}
\label{PW class}}
\subfloat[]{\includegraphics[scale = .72]{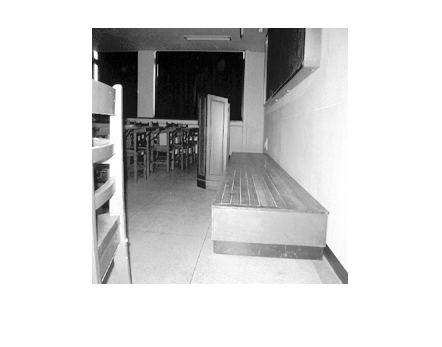}
\label{COW class}}
\hfil
\subfloat[]{\includegraphics[scale = .72]{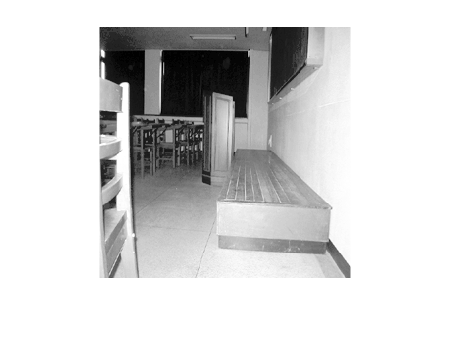}
\label{ar class}}
\subfloat[]{\includegraphics[scale = .72]{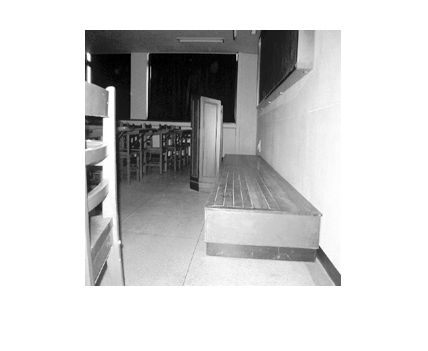}
\label{km class}}
\caption{Enhanced image of \lq{Classroom}\rq for : (a) point-wise, (b) center-of-weights, (c) area, and (d) KM method}
\label{fig_simclass2}
\end{figure}

\newpage
\vspace*{2 cm}

%%%%%%%%%%%%%%%%%%%%%%%%%
\newpage

\begin{figure}[H]
\centering
\subfloat[]{\includegraphics[scale = .4]{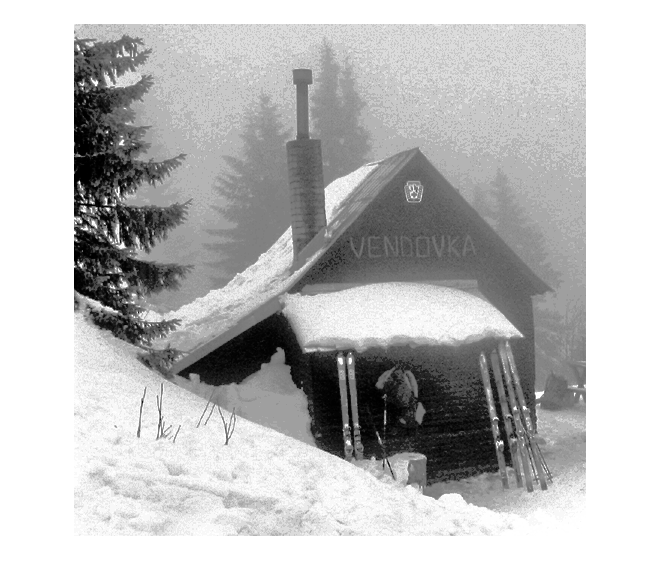}
\label{PW ven}}
\subfloat[]{\includegraphics[scale = .4]{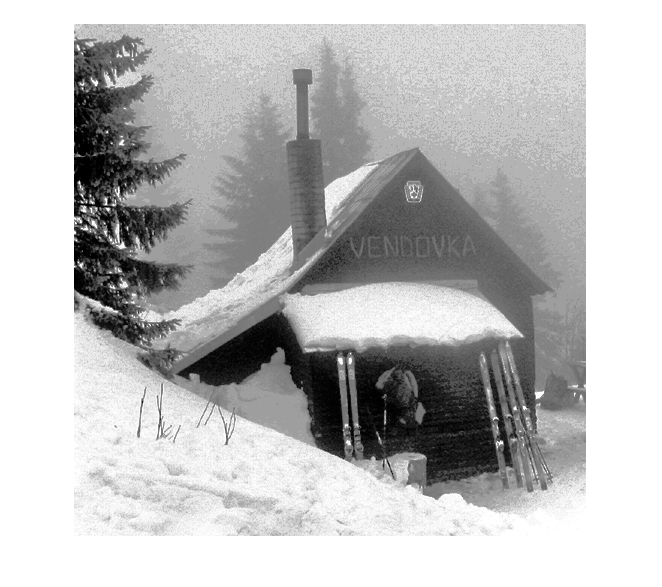}
\label{COW ven}}
\hfil
\subfloat[]{\includegraphics[scale = .4]{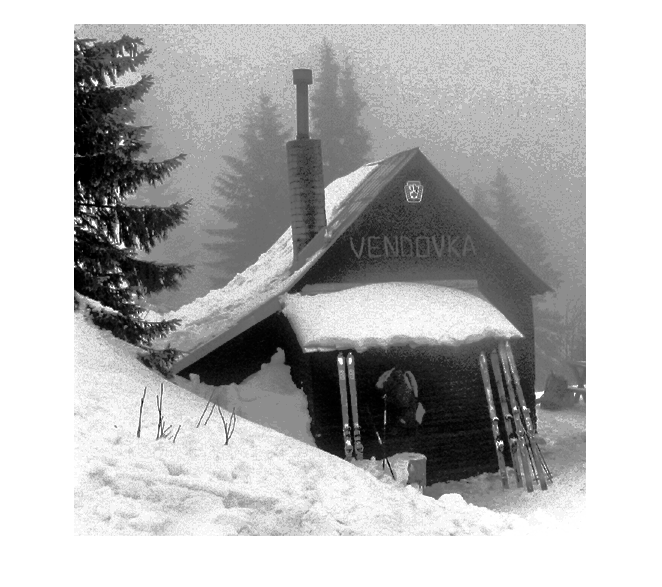}
\label{ar ven}}
\subfloat[]{\includegraphics[scale = .4]{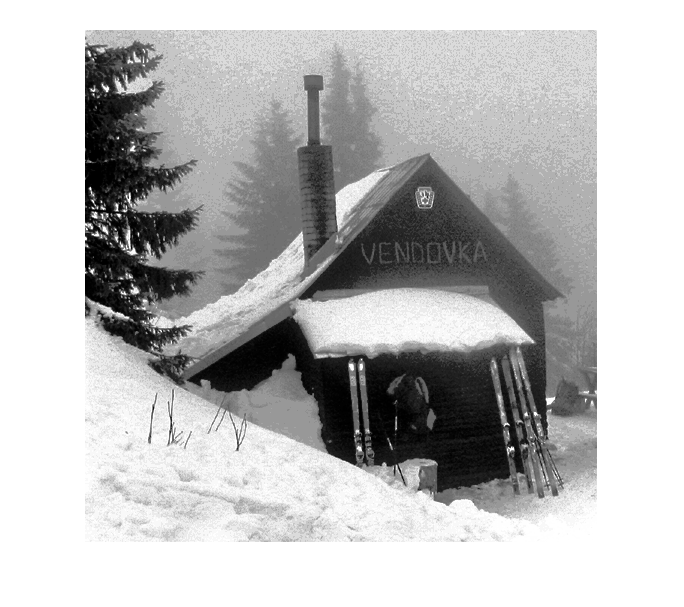}
\label{km ven}}
\caption{Enhanced image of \lq{Vendovka}\rq for : (a) point-wise, (b) center-of-weights, (c) area, and (d) KM method}
\label{fig_simven2}
\end{figure}

\newpage
\vspace*{2 cm}


\clearpage
\section{Conclusion}
Contrast enhancement algorithms aid in extraction of information which may not be easily distinguished in low contrast images. In this paper, we attempted to devise an automatic algorithm which extracts the desired probability distribution function (PDF) for the histogram specification from the input image. The process involved the application of Interval Type-2 fuzzy system and Histogram Specification. The main idea is to spread the histogram of input image in such a way that the image becomes clear and the information content is not lost, which is observed in already present algorithms such as Histogram Equalization(HE), Recursive Mean-Separate Histogram (RMSHE) and Brightness Preserving Fuzzy Histogram Equalization(BPFHE). \par
The proposed algorithm works in 5 stages, each of which address a particular aspect of contrast enhancement. Fuzzy MF extraction replaces the histogram linearisation by fuzzifying discrete gray levels to continuous spectrum, thereby overcoming the limitations imposed by discreteness and generating functional approximation of the histogram.  For getting the footprint of uncertainty, we used the concept of Gaussian fitting, after which we propose 4 methods to get the membership values and hence the desired PDF.  Out of the 4 methods, method of MV extraction using Karnik-Mendel (KM) method is found to be the best for most of the images. Significantly, the proposed algorithm is sensitive to local variations in the histogram and is also modular and flexible (due to the usage of a MV computation techniques). The first stage and MV computation may be computationally expensive but lead to significant improvements in the visual output and in a computer vision pipeline.\par
The proposed algorithm opens up several avenues of further research. For instance, non-linear generation of PDF may improve smoothness in gray level variation of output histogram and preserve image detail. Moreover, We believe that MV computation techniques may be considered to be early attempts for effectively generating general type-2 fuzzy MFs. Thus, an extension of the first stage to higher dimensional fuzzy sets (general type-2) may prove beneficial. Computational complexity may be improved  by optimizing the approximation of histograms or providing an alternate framework for MV computation.\par
The proposed method can also be used as a major component of image enhancement in computer vision pipelines. We believe that applying proposed automatic fuzzy PDF generation for HS into various detection and recognition system, may improve their performance.
\section*{Acknowledgment}

The authors would like to thank Duddu Sai Meher Karthik and Raghav Gulati (Students of Indian Institute of Technology, Guwahati, India)  for giving some important suggestions. The authors thank the anonymous referees for their helpful suggestions for the improvement of the manuscript.



%\vfill


\end{document}